\documentclass{article}
\usepackage[T1]{fontenc}
\usepackage{iclr2026_conference,times}
\iclrfinalcopy

\usepackage{amsmath,amsfonts,bm}

\def\eqref#1{equation~\ref{#1}}
\def\1{\bm{1}}

\DeclareMathAlphabet{\mathsfit}{\encodingdefault}{\sfdefault}{m}{sl}
\SetMathAlphabet{\mathsfit}{bold}{\encodingdefault}{\sfdefault}{bx}{n}

\newcommand{\R}{\mathbb{R}}

\usepackage{hyperref}
\usepackage{url}
\usepackage{booktabs}
\usepackage{amsmath}
\usepackage{graphicx}
\usepackage{capt-of}
\usepackage{xcolor}
\usepackage{algorithm}
\usepackage{algpseudocode}
\usepackage{placeins}
\usepackage{etoc}
\usepackage{arydshln}
\expandafter\let\csname endtabular*\endcsname\endtabular

\def\eqref#1{(\ref{#1})}

\title{VisionWeave: Weaving Elastic Visual Representations as a Native Capability of MLLMs}
\hypersetup{pdftitle={VisionWeave: Weaving Elastic Visual Representations as a Native Capability of MLLMs}}

\makeatletter
\author{\begin{minipage}[t]{\dimexpr\textwidth-2\tabcolsep\relax}
\centering
\def\@makefnmark{\hbox{$^{\@thefnmark}$}}
\fontsize{10}{11.5}\selectfont
\bfseries
Yuan Feng$^{1,2}$\thanks{Equal contribution. $^{\dagger}$Corresponding authors
. Work done as an intern at Qwen Team, Alibaba Token Hub.}, Qize Yang$^{2 \ast}$, Ruizhe Chen$^{2}$, Sibo Song$^{2}$, Haolin He$^{2,3}$, Muzhi Zhu$^{2,4}$,\\
Zihan Liu$^{2}$, Yunfei Chu$^{2}$, Xize Cheng$^{2}$, Yuxuan Wang$^{2}$, Jin Xu$^{2,\dagger}$, Xike Xie$^{1,\dagger}$\\
\normalfont\fontsize{10}{11.5}\selectfont
$^{1}$University of Science and Technology of China\quad
$^{2}$Alibaba Token Hub, Alibaba Group\\
$^{3}$The Chinese University of Hong Kong\quad
$^{4}$Zhejiang University
\end{minipage}}
\makeatother

\newcommand{\method}{VisionWeave}

\begin{document}

\etocdepthtag.toc{main}
\maketitle
\lhead{Preprint}

\begin{abstract}

{

Multimodal large language models have become the dominant paradigm for visual understanding, but incur substantial costs by encoding inputs into dense, fixed-size patch tokens. However, visual information is unevenly distributed: some regions require fine-grained detail, while others admit compact representations. Downsampling sacrifices this detail, while existing token pruning and  adaptive approaches remain limited in content-adaptive granularity, task generalization, and integration with modern MLLMs and serving infrastructure.
Overcoming these limitations calls for foundation models that learn, end to end, where—and at what granularity—to allocate visual representations, a native capability we term \textit{elastic visual representation weaving}. We introduce \textit{VisionWeave}, establishing this capability in frontier-level MLLMs through large-scale training. It combines two components: a gated spatial pooler constructs coarse-grained representations alongside native fine-grained representations within a shared MRoPE coordinate, while a granularity router learns their content-adaptive allocation. Through self-distillation alone, we validate this capability on Qwen3.5-4B and scale to Qwen3.8-27B with over 30K A100 GPU-hours.   Based on Qwen3.8-27B, VisionWeave adaptively adjusts token savings to visual content, saving 43.0\% tokens on average while retaining 98.9\% native performance across eight benchmarks, versus only 88\% performance preserved for token pruning baselines with a fixed 50\% savings target. 
Extensive evaluations confirm robust efficiency–quality trade-offs across diverse tasks, resolutions and video frames. When deployed on SGLang serving engine, our method achieves a 2.30× end-to-end throughput gain while reducing mean TTFT by 54.4\% and mean TPOT by 60.6\%.
Together, we believe these results position elastic visual weaving as a promising capability for next-generation multimodal foundation models,  natively modulating computational effort by information content. Our SGLang code and other details will be available at \url{https://github.com/FFY0/VisionWeave}. 
}
  
 \end{abstract}

\begin{minipage}[t]{\linewidth}
  \centering
  \vspace{-0.5cm}
  \includegraphics[width=0.95\linewidth]{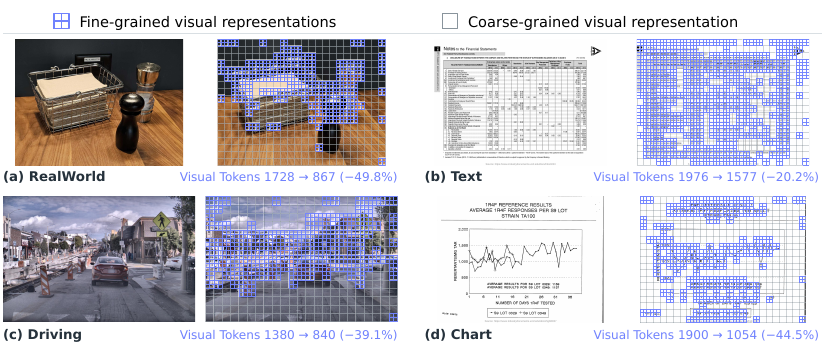}
  %\begingroup
  \small
  \vspace{-0.3cm}
  \captionof{figure}{Weaving elastic visual representations.
   More examples appear in
  Appendix~\ref{sec:more_visualizations}.}
  \label{fig:router-teaser}
 % \endgroup
\end{minipage}

\section{Introduction}

Multimodal large language models (MLLMs) have become the dominant paradigm for visual understanding, supporting diverse tasks such as visual question answering, document analysis, and video understanding within a unified framework~\citep{Bai2025Qwen25VLTR, Bai2025Qwen3VLTR}. As these tasks demand finer spatial detail and broader temporal coverage, MLLMs must process higher-resolution images and longer sequences of video frames. However, their visual inputs are typically partitioned into fixed-size patches and encoded into dense sequences of visual tokens. The resulting growth in sequence length places substantial computational and memory demands, making visual efficiency a central challenge in scaling multimodal capabilities~\citep{Shao2025ASO,Tao2024DyCokeD,Shao2025WhenTT}.

  Visual downsampling offers a straightforward solution but sacrifices
  fine-grained detail, motivating extensive research into visual token
  pruning and merging~\citep{Chen2024AnII,Yang2024VisionZipLI,Ye2024ATPLLaVAAT,guo2026videommadaptivemacromicroinference}.
  However, both heuristic and learned methods introduce a training–inference mismatch and discard task-specific information. Our evaluations
  of FastV and VisionZip reveal severe degradation on
  information-dense inputs and grounding tasks. Common design choices in these methods,
  including attention-weight extraction and in-LLM pruning, also complicate
  integration with modern serving engines such as SGLang and vLLM,
  where implementations remain rare and practical efficiency gains
  insufficiently validated. Preset compression ratios further limit
  adaptivity to diverse visual inputs. These limitations motivate native,
  content-adaptive visual processing for general-purpose MLLMs:
  learning where to retain fine-grained detail and where---and to what
  extent---to use compact coarse representations.

Recent work has explored content-adaptive visual representations, but
  realizing this adaptivity as a native capability of general-purpose MLLMs
  remains challenging. One line of work varies ViT patch sizes using
  low-level image statistics, such as edge density and
  entropy~\citep{Yu2025DynamicGM,Choudhury2025AcceleratingVT}.
  These methods primarily target standalone vision tasks, and their
  granularity criteria are not optimized for diverse downstream multimodal
  objectives. ViCO~\citep{Cui2025ViCOAT} is particularly close to our approach:
  it partitions an image into tiles and adaptively assigns each a visual
  token resolution. However, its tile-based design limits 
  applicability to modern MLLMs such as Qwen~\citep{Bai2025Qwen3VLTR} and Kimi~\citep{Bai2026KimiKO}, which use
  native-resolution visual processing. Moreover, detail-rich regions and
  low-information backgrounds are often interleaved
  (Figure~\ref{fig:router-teaser}), requiring finer-grained spatial allocation
  than a single resolution per tile can provide.\footnote{See
  related work section for further discussion in Appendix~\ref{app:related} and Figure~\ref{fig:app-vico-granularity} for a visual comparison.}

\begin{figure}[t!]
  \centering
    \vspace{-0.2cm}
  \includegraphics[width=\linewidth]{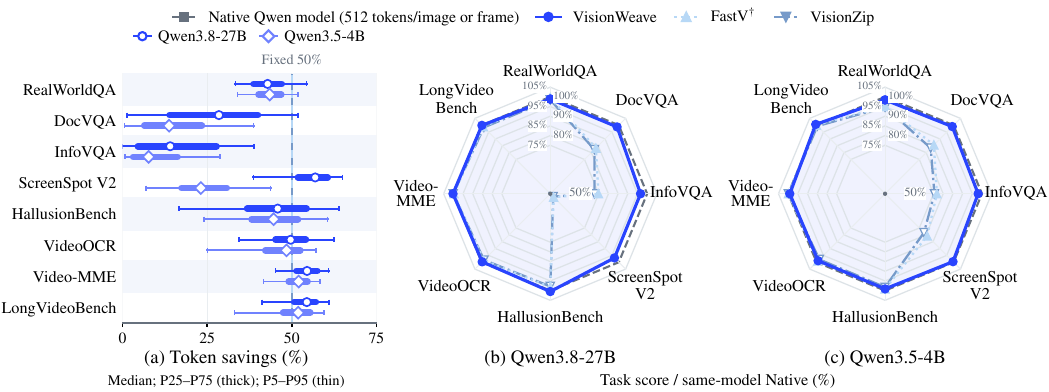}
  \setlength{\abovecaptionskip}{2pt}
  \setlength{\belowcaptionskip}{0pt}
  \vspace{-0.1cm}
  \caption{\method{} achieves content-adaptive token savings with near-native quality, whereas fix-budget baselines degrade sharply on challenging tasks. Visual case studies on ScreenSpotV2 and DocVQA illustrate this difference (Appendix~\ref{sec:app-case-studies}).
  Our \method{} still retains higher quality even when baselines are granted its adaptive saving ratios (Appendix~\ref{sec:exp-matched}). }
  \label{fig:teaser}
  \label{fig:adaptive-eightbench-512}
\end{figure}

\begingroup
\setlength{\fboxsep}{7pt}
\noindent
\colorbox{black!5}{%
\begin{minipage}{\dimexpr\linewidth-2\fboxsep\relax}
\setlength{\parskip}{3pt}
\itshape
We envision elastic visual representation weaving as a native capability of foundamental MLLMs that should satisfy three requirements:

\par\noindent
1. Autonomously content-adaptive token savings while preserving quality.
\par\noindent
2. Consistently favorable trade-off than downsampling across diverse visual task inputs.
\par\noindent
3. Compatibility with frontier-level MLLM architectures and modern serving infrastructure.
\end{minipage}
}
\par
\endgroup

To this end, we introduce VisionWeave, which enables MLLMs to operate natively on elastic visual representations, adaptively weaving together fine- and coarse-grained representations according to visual content. To our knowledge, this is the first effort to establish elastic visual representation weaving as a native capability of frontier MLLMs through large-scale training, accompanied by extensive evaluation to assess its viability as an integral capability of foundation models.
The method comprises two core components. First, the elastic representation itself: a gated spatial pooler constructs a learned coarse-grained representation that complements the native fine-grained tokens within the same MRoPE frame. Second, learned granularity allocation: a granularity router probes ViT features to determine the appropriate granularity for each spatial block.
Through self-distillation alone, we make elastic visual processing a native capability, validating on Qwen3.5-4B and then scaling to frontier Qwen3.8-27B with over 30K A100 GPU-hours.  

With this native capability, VisionWeave preserves fine-grained detail where needed and uses compact representations elsewhere (Figure~\ref{fig:router-teaser}), achieving content-adaptive token savings with minimal impact on task quality (Figure~\ref{fig:teaser}). At an input budget of 512 tokens per image, VisionWeave on Qwen3.8-27B saves 16.5--55.1\% of visual tokens across eight benchmarks, averaging 43.0\% savings while preserving 98.9\% native performance. In contrast, FastV$^\dagger$ and VisionZip retain only 88.2\% and 87.5\%, respectively, at a fixed 50\% savings target. Broader evaluations across input resolutions and video frame budgets demonstrate consistently more favorable efficiency--quality trade-offs than input downsampling. Crucially, these token savings translate into practical serving gains build upon modern SGLang serving engine, achieving  around $2\times$ throughput improvements. Together, content-adaptive savings, cross-task robustness, and practical deployability support elastic visual processing as a native capability of next-general efficient MLLMs.

\section{Method}

In this work, we realize elastic visual representation weaving through two complementary
components. A {gated spatial pooler} constructs learned coarse-grained
representations that complement native fine-grained tokens (\S\ref{sec:rep}). A {granularity router} probes intermediate
ViT features to determine the granularity for each spatial block
  (\S\ref{sec:router}). Through self-distillation alone, we make elastic
  visual processing a native capability of MLLMs (\S\ref{sec:train}).

\begin{figure}[t]
\centering
\includegraphics[width=0.9\linewidth]{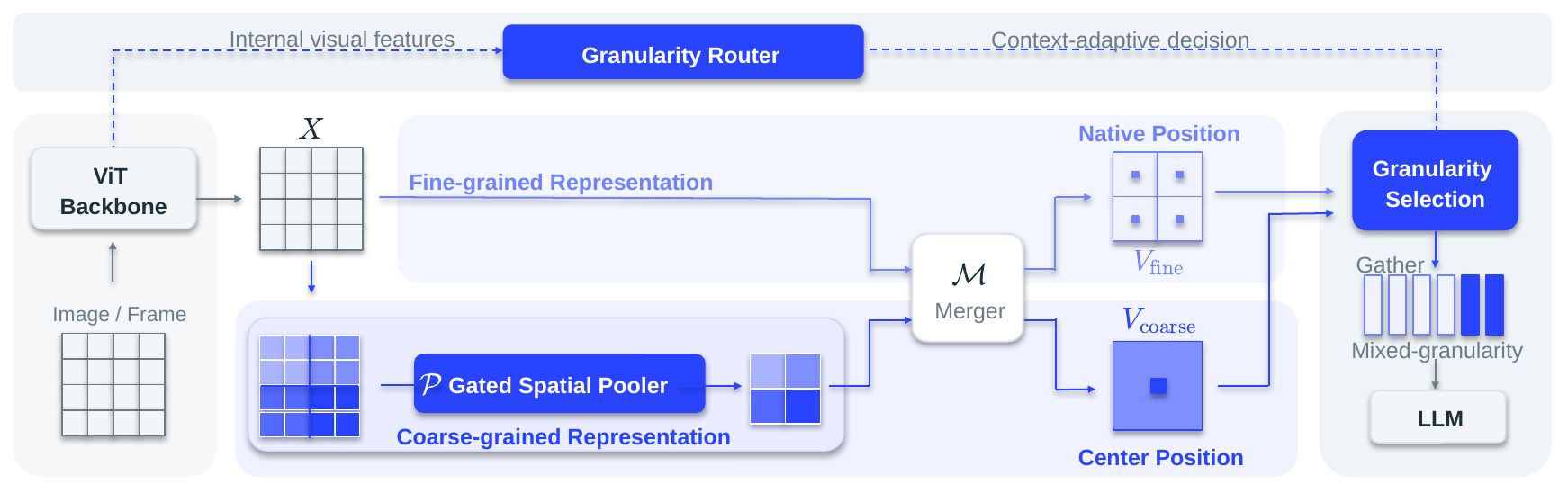}
\vspace{-0.2cm}
\caption{Overall architecture of \method{}.
The gated spatial pooler complements native fine-grained tokens $V_{\mathrm{fine}}$ with coarse-grained representations $V_{\mathrm{coarse}}$ within the same
MRoPE coordinate, while the granularity router
adaptively weaves them into a mixed-granularity sequence for the LLM.}
\vspace{-0.2cm}
\label{fig:pipeline}
\end{figure}

\subsection{Preliminaries}
\label{sec:prelim}

For modern MLLMs, the ViT produces patch features
$X\in\R^{H\times W\times d}$. A native merger $\mathcal{M}$
concatenates each $2{\times}2$ neighborhood and applies an MLP to
produce  fine-grained visual tokens:
\begin{equation}
\small
V_{\mathrm{fine}}=\mathcal{M}(X)
\in\R^{\frac{H}{2}\times\frac{W}{2}\times D},
\label{eq:native}
\end{equation}
where $d$ and $D$ are the ViT and LLM hidden sizes, respectively.
MRoPE assigns each token coordinates $\pi_{ij}=(\tau,i,j)$,
with temporal coordinate $\tau$ and spatial indices $(i,j)$. The flattened visual tokens and textual context
$c$ condition autoregressive response generation,
$p_\theta(y\mid c,V_{\mathrm{fine}})$.
% Throughout, $\operatorname{flatten}$ denotes grid-to-sequence
% conversion, which is implicit in LLM conditioning arguments.

\subsection{Architecture}

\subsubsection{Gated Spatial Pooler}
\label{sec:rep}

{
Pooling adjacent patch features is a natural compression primitive: the patches
 of a block cover contiguous regions whose content is typically
 correlated---increasingly so wherever the image is locally homogeneous. Leveraging this correlation, Our \emph{gated spatial pooler} $\mathcal{P}$ maps each $2{\times}2$ neighborhood of ViT patch features to a single pooled feature. Passing these features through the native merger yields the coarse-grained representation in the same embedding space.
  \begin{equation}
  \small
  V_{\mathrm{coarse}}=\mathcal{M}\big(\mathcal{P}(X)\big)\in\R^{\frac{H}{4}\times\frac{W}{4}\times D},
  \qquad \mathcal{P}(X)\in\R^{\frac{H}{2}\times\frac{W}{2}\times d}.
  \label{eq:compressed}
  \end{equation}
  Each block $(I,J)$ on the $\tfrac{H}{4}{\times}\tfrac{W}{4}$ grid
  covers $4{\times}4$ patches and admits either four native fine-grained
  tokens $\{V_{\mathrm{fine},2I+a,2J+b}\}_{a,b\in\{0,1\}}$
  or one coarse-grained token $V_{\mathrm{coarse},IJ}$. We use positional interpolation~\citep{Peng2023YaRNEC, Chen2023ExtendingCW} to place each coarse token at the geometric center
  of its four corresponding fine-grained tokens. 
 % Such positional interpolation is a widely used technique with RoPE, and prior work has shown its generalization [CITE].
  \begin{equation}
  \pi_{IJ}=\big(\tau,\; 2I+\tfrac{1}{2},\; 2J+\tfrac{1}{2}\big).
  \label{eq:center}
  \end{equation}
}

We implement the gated spatial pooler $\mathcal{P}$ as a learnable
  extension of mean pooling, combining content-dependent and channel-wise weighting with
  feature refinement. For each $2{\times}2$ neighborhood, we normalize
  its four patch features $\{x_k\}_{k=1}^{4}$ using learnable layer
  normalization~\footnote{ViT patch features exhibit large magnitude variation; LayerNorm prevents high-magnitude tokens from dominating the pooling output.} and aggregate them into a shared global context $g_k$:
\begin{equation}
u_k = \mathrm{LN}(x_k) \in \R^{d}, \qquad
g_k = f_{\mathrm{glb}}\big([u_1;u_2;u_3;u_4]\big) + b_k \;\in \R^{d},
\label{eq:gsp-summary}
\end{equation}
where $[\cdot;\cdot]$ denotes concatenation and $b_k\in\R^d$
  is a learnable position-specific bias~\citep{DeepSeekAI2026DeepSeekV4TH}.
  The gate then computes channel-wise scores from patch feature
  $u_k$ and its context $g_k$:
% where $[u_1;u_2;u_3;u_4] \in \R^{4d}$ concatenates the four normalized
% features, and $b_k \in \R^{d}$ is a learnable bias marking corresponding 
% positions~\cite{DeepSeekAI2026DeepSeekV4TH}. A gate then scores each
% position from the patch itself together with this context,
\begin{equation}
\alpha_k = \operatorname*{softmax}_{k=1,\dots,4}
f_{\mathrm{gate}}\big([u_k;\, g_k]\big) \;\in (0,1)^{d},
\label{eq:gsp-gate}
\end{equation}
where the softmax normalizes {over the four spatial positions}
channel-wise, i.e.\ $\sum_{k=1}^{4}\alpha_k = \mathbf{1}_d$. Finally, a value
net $f_{\mathrm{val}}$ refines each feature, and the pooled output is the
gated sum,
\begin{equation}
v_k = u_k + f_{\mathrm{val}}\big([u_k;\, g_k]\big) \;\in \R^{d}, \qquad
\mathcal{P}(x_{1:4}) = \sum\nolimits_{k=1}^{4}\alpha_k\odot v_k \;\in \R^{d},
\label{eq:gsp-out}
\end{equation}
where $f_{\mathrm{glb}}$, $f_{\mathrm{gate}}$, $f_{\mathrm{val}}$ are
two-layer MLPs (each mapping into $\R^{d}$) and $\odot$ is the elementwise
product.  Beyond its greate expressiveness~\footnote{Ablation study in Section~\ref{sec:exp-pooler-ablation} shows gated pooler consistently outperforms the widely used pixel-unshuffle.}, the pooler offers a favorable starting point: zero-initializing the final layers of $f_{\mathrm{gate}}$ and $f_{\mathrm{val}}$ reduces $\mathcal{P}$ to mean pooling over normalized patch features.

\subsubsection{Granularity Router}
\label{sec:router}

{
With both granularities available, our \emph{granularity router} learns
  their content-adaptive allocation across spatial blocks \footnote{Ablation studies in Section~\ref{sec:exp-router-ablation} validate the effectiveness of the learned router.}. It probes ViT
  features at multiple depths with a learnable query, combining low-level
  detail from early layers with semantic information from later layers.  We implement this multi-depth probing with one query state
  $h_{IJ}\in\R^{d}$ per spatial block, initialized from a shared learnable
  vector. Each query uses the block-center MRoPE coordinate
  $\big(4I+\tfrac{3}{2},\,4J+\tfrac{3}{2}\big)$ on the
  $H{\times}W$ patch grid, corresponding to the same spatial center as
  Eq.~\eqref{eq:center}. For a ViT with $L$ layers, we denote the features
  after layer $\ell$ by $X^{(\ell)}\in\R^{H\times W\times d}$,
  with $X^{(0)}$ the patch embeddings and $X^{(L)}=X$. Each query state is updated through cross-attention to ViT features, followed by an MLP, both with residual connections:
}
\begin{equation}
\small
h'=\mathrm{LN}\Big(h+\mathrm{CrossAttn}\big(h,\;\mathrm{LN}(X^{(\ell)})\big)\Big),
\qquad
h\leftarrow\mathrm{LN}\big(h'+\mathrm{MLP}(h')\big),
\label{eq:router-block}
\end{equation}
where $\mathrm{CrossAttn}$ takes its query from $h$ and keys/values from
$\mathrm{LN}(X^{(\ell)})$, with MRoPE applied to queries and keys; the KV-side
$\mathrm{LN}$ keeps features tapped at different depths on a comparable
scale.  Each query sequentially probes depths $\ell=0,\,L//2,\,L$.
  At each depth, a local block is followed by a global block, yielding
  six independently parameterized blocks in total. The local block attends
  to the spatial block's own $4{\times}4$ patch features
  $X^{(\ell)}_{IJ}\in\R^{4\times4\times d}$; a \emph{global} block then opens them to the whole frame:
\begin{equation}
\small
h_{IJ}\leftarrow
\mathrm{LocalCrossAttn}^{(\ell)}\big(h_{IJ},\,
{X^{(\ell)}_{IJ}}\big);\: \: h_{IJ}\leftarrow
\mathrm{GlobalCrossAttn}^{(\ell)}\big(h_{IJ},\,
{X^{(\ell)}}\big).
\label{eq:router}
\end{equation}
  Finally, a linear head maps the normalized query state to two logits,
  giving the probability of using the coarse-grained representation:
\begin{equation}
p_{IJ}=\operatorname{softmax}\big(W_{\!r}\,\mathrm{LN}(h_{IJ})\big)_{\mathrm{coarse}},
\qquad W_{\!r}\in\R^{2\times d} .
\label{eq:router-head}
\end{equation}
For each spatial block, we use its coarse-grained token if the routing probability
  exceeds a threshold \footnote{We set $t=0.5$ by default, equivalent to selecting the granularity
  with the higher predicted probability. Section~\ref{sec:exp-control}
  further shows how varying this threshold controls the compression level
  at inference time.} and retain its four native fine-grained tokens otherwise.
  The selected tokens are gathered into a mixed-granularity sequence for
  the LLM backbone within the shared MRoPE frame.
  Algorithm~\ref{alg:inference} 
  summarizes the complete inference path.

\begin{algorithm}[!t]
\caption{\method{} inference with elastic visual representations}
\label{alg:inference}
\begin{algorithmic}[1]
\Require An $H{\times}W$ patch grid for an image; textual context $c$;
routing threshold $t$ (default $0.5$)
\State Obtain $X^{(0)},\,X^{(L/2)},\,X^{(L)}$ in a single ViT forward pass
\State $V_{\mathrm{fine}}\leftarrow\mathcal{M}(X^{(L)})$;\quad
$V_{\mathrm{coarse}}\leftarrow\mathcal{M}\big(\mathcal{P}(X^{(L)})\big)$
\hfill Eqs.~\eqref{eq:native},~\eqref{eq:compressed}
\State Initialize $h_{IJ}$ from the shared query with block-center RoPE coordinates
\For{$\ell\in\{0,\,L//2,\,L\}$}
  \State $h_{IJ}\leftarrow
  \mathrm{LocalCrossAttn}^{(\ell)}\big(h_{IJ},\,X^{(\ell)}_{IJ}\big)$ 
  \State $h_{IJ}\leftarrow
  \mathrm{GlobalCrossAttn}^{(\ell)}\big(h_{IJ},\,X^{(\ell)}\big)$ 
  \hfill Eq.~\eqref{eq:router}
\EndFor
\State Compute $p_{IJ}$ using the calibrated router head
\For{each block $(I,J)$}
  \If{$p_{IJ} > t$}
    \State Select $V_{\mathrm{coarse},IJ}$ with its block-center coordinate
    $\pi_{IJ}$ from Eq.~\eqref{eq:center}
  \Else
    \State Select $\{V_{\mathrm{fine},2I+a,2J+b}\}_{a,b\in\{0,1\}}$
    with their native coordinates from Section~\ref{sec:prelim}
  \EndIf
\EndFor
\State Gather selected tokens into $V_{\mathrm{routed}}$ in native raster order
\State \Return $p_\theta\big(y\mid c,\,V_{\mathrm{routed}}\big)$
\end{algorithmic}
\end{algorithm}

\subsection{Training}
\label{sec:train}

We develop elastic visual weaving through three-stage self-distillation
  (Table~\ref{tab:stages}).\footnote{Section~\ref{sec:training-details}
  details the training data and settings and analyzes loss and routing
  dynamics.} The native forward pass provides output-distribution
  supervision. Distillation is off-policy \footnote{We try switching to on-policy self-distillation in Stage~3
  of Qwen3.5-4B but observe no consistent gains
  (Appendix~\ref{sec:app-policy-ablation}). We hypothesize that elastic
  visual weaving keeps the student's output distribution close to the
  teacher's, limiting the benefit of on-policy sampling; we therefore
  use off-policy training for efficiency.}, with teacher and student
  conditioned on the same teacher rollout.
  The student progresses from coarse-grained representations to a
  differentiable soft-mixing surrogate, and finally to the hard-routed
  mixed-granularity sequences used at inference.
\begin{table}[t]
\vspace{-0.2cm}
\caption{Three-stage self-distillation for elastic visual processing.}
\label{tab:stages}
\begin{center}
\small
\setlength{\tabcolsep}{4pt}
\begin{tabular}{@{}llll@{}}
\toprule
 & Stage 1 & Stage 2 & Stage 3 \\
\midrule
Trainable components & Gated Spatial Pooler & Granularity Router & Pooler, LLM \\
Teacher       & Native model       & Native model                & Native model \\
Routing       & Bypassed (all coarse) & Soft-mixing routing      & Hard routing \\
Loss terms    & $\mathcal{L}_{\mathrm{distill}}$ & $\mathcal{L}_{\mathrm{distill}}+0.02\mathcal{L}_{\mathrm{bal}}$ & $\mathcal{L}_{\mathrm{distill}}$ \\
Student visual tokens & $25\%$  & $100\%$      & Content-adaptive \\
\bottomrule
\end{tabular}
\end{center}
\vspace{-0.2cm}
\end{table}

\subsubsection{Stage 1: Coarse-Grained Self-Distillation}
\label{sec:stage1}

 We first train the pooler with routing bypassed, using the coarse-grained
  representation for every block:
  $V_{\mathrm{student}}=\operatorname{flatten}(V_{\mathrm{coarse}})$.
  All other components remain frozen. The model's native forward pass
  serves as the teacher, yielding the distillation objective:
  \begin{equation}
  \small
  \mathcal{L}_{\mathrm{distill}}
  =\frac{1}{|y|}\sum_{t}
  \mathrm{KL}\Big(
  \widehat{p}_{\bar\theta}\big(\cdot \mid y_{<t},\,c,\,V_{\mathrm{fine}}\big)
  \;\Big\|\;
  \widehat{p}_{\theta}\big(\cdot \mid y_{<t},\,c,\,V_{\mathrm{student}}\big)\Big),
  \label{eq:distill}
  \end{equation}
  where $y$ is a response generated by the teacher and $\bar\theta$ denotes
  its parameters. Both $\widehat{p}_{\bar\theta}$ and $\widehat{p}_{\theta}$
  are renormalized over the teacher's top-$512$ vocabulary entries at each
  response position.
  Later stages reuse this objective with different constructions of
  $V_{\mathrm{student}}$.

\subsubsection{Stage 2: Soft-Mixing Self-Distillation}
\label{sec:stage2}

We next train only the granularity router using
a differentiable soft-mixing surrogate for hard routing. We interpolate
feature directions and magnitudes separately to prevent norm differences
from biasing the directional mixture. Let $\widehat{V}_{\mathrm{fine}}$
and $\widehat{V}_{\mathrm{coarse}}$ denote the token-wise L2-normalized
representations under $\mathcal{N}(v)=v/\max(\|v\|_2,\epsilon)$.
  We compute
 \begin{equation}
 \small
  \begin{aligned}
  m_{ij}
    &=(1-p_{IJ})\|V_{\mathrm{fine},ij}\|_2
      +p_{IJ}\|V_{\mathrm{coarse},IJ}\|_2,\\
  V_{\mathrm{mix},ij}
    &=m_{ij}\,\mathcal{N}\!\left(
      (1-p_{IJ})\widehat{V}_{\mathrm{fine},ij}
      +p_{IJ}\widehat{V}_{\mathrm{coarse},IJ}
      \right).
  \end{aligned}
  \label{eq:soft-mix}
\end{equation}
  where $I=\lfloor i/2\rfloor$, $J=\lfloor j/2\rfloor$, and
  $\epsilon$ ensures numerical stability. Each coarse token is broadcast
  to its four corresponding native positions. The resulting sequence,
  $V_{\mathrm{student}}=\operatorname{flatten}(V_{\mathrm{mix}})$,
  is used in the distillation objective of Eq.~\eqref{eq:distill}.
  We add a per-sample balance loss borrow from MoE training:
  \begin{equation}
  \small
  \mathcal{L}_{\mathrm{bal}}
    = \frac{{F}}{\rho}\,G
    + \frac{1-{F}}{1-\rho}\,(1-G),
  \label{eq:balance}
  \end{equation}
  where $F$ is the fraction of the sample's blocks with $p_{IJ}\ge0.5$,
  $G$ is its mean routing probability, $\rho$ is the target coarse-routing
  fraction. The penalty encourages more coarse routing when $F<\rho$ and less
  when $F>\rho$. Per-sample statistics prevent larger inputs from
  dominating the estimate and batch averages from masking input-level
  imbalance. We set $\rho=0.8$ and optimize
  $\mathcal{L}_{\mathrm{distill}}+0.02\mathcal{L}_{\mathrm{bal}}$.
  % This target corresponds to retaining $40\%$ of native visual tokens
  % under hard routing, but serves only to regularize router warm-up,
  % not to prescribe an inference-time token budget.

\subsubsection{Stage 3: Mixed-Granularity Self-Distillation}
\label{sec:stage3}

Stage~3 performs self-distillation on the hard-routed mixed-granularity
sequences used at inference. \footnote{Appendix~\ref{sec:app-stage3-ablation} shows that Stage~3 improves performance across diverse tasks, especially on grounding.}
The router head includes an offline-calibrated scalar coarse-routing bias,
held fixed during training and inference, without auxiliary balancing losses
or online adjustments. \footnote{In experiments, we calibrate on a training subset targeting 50\% visual-token savings.} The student gathers $V_{\mathrm{student}}$ from
$V_{\mathrm{fine}}$ and $V_{\mathrm{coarse}}$ using the same token
selection and RoPE coordinates as at inference. The native forward
pass of a frozen teacher provides supervision through the
distillation objective.

\subsection{Training Details and Dynamics}
\label{sec:training-details}
\label{sec:app-training}
\label{sec:training-dynamics}

\paragraph{Training data.}
All three stages of Qwen3.8-27B training use the same mixture of
approximately $0.78$ million samples covering image understanding,
video understanding, and GUI grounding
(Table~\ref{tab:training-data}). Counts refer to samples before
sequence packing. The visual input budgets are $4{,}096$ native visual
tokens per image and $512$ per video frame, with up to $64$ sampled frames.

\begin{table}[!t]
\caption{Training-data composition for Qwen3.8-27B.}
\label{tab:training-data}
\centering
\small
\begin{tabular}{@{}llr@{}}
\toprule
Source & Domain & Approx. training samples \\
\midrule
LLaVA-OneVision~\citep{an2025llavaonevision15fullyopenframework} & Image understanding & 246K \\
LLaVA-Video-178K~\citep{zhang2025llavavideovideoinstructiontuning} & Video understanding & 281K \\
UGround~\citep{gou2025navigatingdigitalworldhumans} & GUI grounding & 250K \\
\midrule
\textbf{Total} & & \textbf{777K} \\
\bottomrule
\end{tabular}
\end{table}

\paragraph{Training settings.}
The granularity router samples ViT features at depths $[0,13,27]$,
with one local--global cross-attention pair per depth. Its six layers
use $16$ attention heads, an MLP expansion ratio of $2$, and key/value
layer normalization; the scoring head is kept in FP32. All stages
use BF16 training with Adam ($\beta_1=0.9$, $\beta_2=0.999$,
$\epsilon=10^{-8}$), zero weight decay and cosine learning-rate decay. Table~\ref{tab:app-training-config}
summarizes the stage-specific settings. Distillation is off-policy
at temperature $1$, using forward KL divergence over  top-$512$ vocabulary entries. Stage~1 weights each sample's
mean token loss by the square root of its supervised response length.
We observed training instability with this weighting in Stage~2 and
therefore use a global mean over supervised response tokens in
Stages~2 and~3. Stage~3 updates the pooler and LLM, while keeping the
ViT, native merger, and router frozen. Global batch sizes count packed
sequences in Stages~1 and~2 and individual samples in Stage~3.

\begin{table}[htbp]
\caption{Training hyperparameters for Qwen3.8-27B.}
\label{tab:app-training-config}
\centering
\small
\begin{tabular}{@{}lccc@{}}
\toprule
Setting & Stage 1 & Stage 2 & Stage 3 \\
\midrule
Peak learning rate & $1 \times 10^{-4}$ & $1 \times 10^{-5}$ & $3{\times}10^{-6}$ \\
Minimum learning rate & $5{\times}10^{-6}$ & $0$ & $0$ \\
Warmup fraction & $5\%$ & $5\%$ & $2\%$ \\
Global batch size & $32$ & $64$ & $512$ \\
Training steps & $5{,}257$ & $2{,}500$ & $1{,}000$ \\
Gradient clipping & $1.0$ & $1.0$ & $0.5$ \\
Student maximum sequence length & $16{,}384$ & $24{,}576$ & $24{,}576$ \\
Packed sequence length & $16{,}384$ & $24{,}576$ & Disabled \\
\bottomrule
\end{tabular}
\end{table}

\begin{figure}[!t]
\centering
\includegraphics[width=\linewidth]{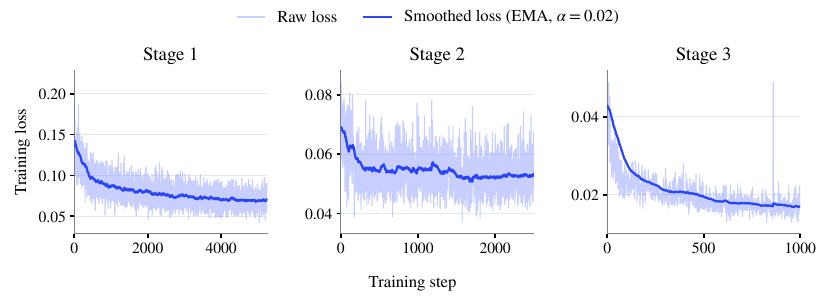}
\par\smallskip
\includegraphics[width=0.8\linewidth]{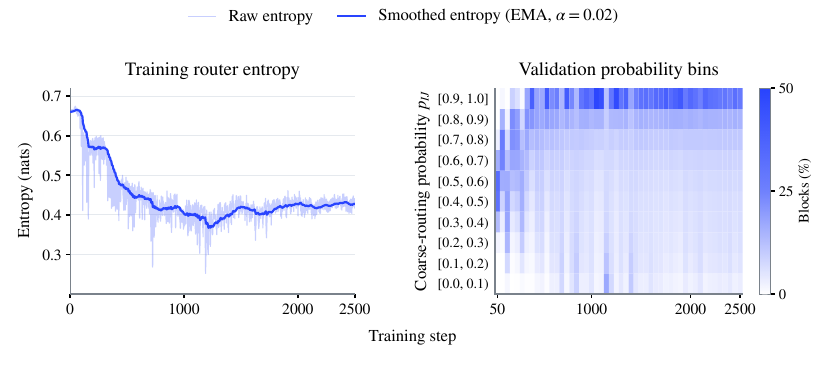}
\vspace{-0.3cm}
\caption{Training dynamics of three-stage self-distillation.}
\label{fig:training-dynamics}
\end{figure}

\paragraph{Training dynamics.}
Figure~\ref{fig:training-dynamics} traces the three self-distillation
stages on Qwen3.8-27B, showing raw losses and exponential moving
averages (EMA, $\alpha=0.02$, initialized at the first step). The routing distributions reveal how granularity decisions develop.
In Stage~2, mean router entropy falls from $0.658$ to $0.427$ nats
between the first and last 100 steps. From step~50 to~$2{,}500$,
the validation share of blocks with $p_{IJ}\in[0.4,0.6)$ decreases
from $64.8\%$ to $14.6\%$, while the share with $p_{IJ}\ge0.9$
reaches $28.3\%$. These changes indicate increasingly decisive
allocation; Section~\ref{sec:analysis-ablation} tests its contribution
to task quality.

\subsection{Deploy \method{} on Modern Serving Infrastructure}
\label{sec:serving}

To demonstrate how native elastic visual weaving can be deployed on modern serving infrastructure, we integrate VisionWeave into SGLang~\citep{Zheng2023SGLangEE}.  This requires addressing two core challenges: supporting block-center MRoPE coordinates and mixed-granularity visual sequences.

\paragraph{Block-center RoPE support.}
Coarse-grained tokens use the block-center RoPE coordinates defined in
Eq.~\eqref{eq:center}. To support these half-integer positions in
SGLang's integer-indexed RoPE caches, we double all position coordinates
and halve rotary frequencies, preserving rotary phases. We apply this
rescaling to cache construction and growth and to position updates during
prefill and decode, reusing the existing rotary-embedding and attention
kernels.

\paragraph{Mixed-granularity sequence support.}
We extend SGLang’s encoder-disaggregated stage by attaching layout metadata to visual responses. This enables pre-admission embedding slicing and RoPE construction, finalizing sequence layouts before scheduling while preserving dynamic batching, KV-cache management, and CUDA-graph decoding. Serving gains are reported in Section~\ref{sec:exp-serving}.
\FloatBarrier
% Main manuscript source; edit this file directly after the Demo integration.
\section{Experiments}
\label{sec:exp}

We organize our evaluation around the three requirements for elastic
  visual weaving as a native MLLM capability: quality-preserving,
  content-adaptive token savings (Section~\ref{sec:exp-main}), robust
  trade-offs  across
  resolutions and frame budgets (Section~\ref{sec:exp-resolution}),
  and practical serving gains on SGLang (Section~\ref{sec:exp-serving}).
  We then examine inference-time control (Section~\ref{sec:exp-control})
  and analyze component contributions and granularity allocation
  (Section~\ref{sec:analysis}).

\subsection{Experimental Setup}
\label{sec:exp-setup}

\paragraph{Models and baselines.}
 We evaluate with two frontier open-source MLLMs, the Qwen3.5-4B and Qwen3.8-27B.  We compare VisionWeave against two heuristic visual-token reduction methods:
  VisionZip~\cite{Yang2024VisionZipLI} and FastV~\cite{Chen2024AnII},
  representing token merging and pruning, respectively.
  To enable a consistent evaluation within SGLang and support its
  modern serving features, we use a FastV variant, denoted
  FastV$^\dagger$, that selects tokens after the vision encoder
  rather than within the LLM.
  Unless otherwise stated, \method{} uses a default routing threshold of $t=0.5$.

\paragraph{Tasks.}
We assess cross-domain robustness on eight benchmarks spanning
natural-image (RealWorldQA), document and infographic
understanding (DocVQA and InfoVQA), GUI grounding (ScreenSpotV2),
visual hallucination (HallusionBench), video details (VideoOCR), and video understanding (Video-MME and
LongVideoBench). All task scores are reported on a 0--100 scale, using ANLS for DocVQA
and InfoVQA, aAcc for HallusionBench, and the task-specific metric
for each remaining benchmark. 

\subsection{Content-Adaptive Token Savings with Preserved Quality}
\label{sec:exp-main}

\begin{table}[!t]
\caption{{Content-adaptive token savings with near-native quality.}}
\label{tab:exp-main}
\centering
\fontsize{8}{9.5}\selectfont
\setlength{\tabcolsep}{2.5pt}
\renewcommand{\arraystretch}{1.05}
\setlength{\dashlinedash}{1.2pt}
\setlength{\dashlinegap}{2pt}
\setlength{\arrayrulewidth}{0.3pt}
\begin{tabular*}{\linewidth}{@{\extracolsep{\fill}}lr:rrrrr@{}}
\toprule
& \multicolumn{1}{c}{Native Model} & \multicolumn{3}{:c}{Fixed Savings (s=50\%)} & \multicolumn{2}{c@{}}{Content-adaptive Savings} \\
\cmidrule(lr){2-2}\cmidrule(lr){3-5}\cmidrule(l){6-7}
\textbf{Qwen3.5-4B} & 512 token/Image & Downsampling & \shortstack{FastV$^\dagger$} & \shortstack{VisionZip} & \shortstack{\method{}} & \shortstack{Savings (\%)} \\
\midrule
RealWorldQA & 74.64 & \underline{72.16}\hspace{1pt}{\fontsize{7}{8}\selectfont\textcolor[gray]{0.35}{\makebox[30pt][r]{($\downarrow$3.32\%)}}} & 70.98\hspace{1pt}{\fontsize{7}{8}\selectfont\textcolor[gray]{0.35}{\makebox[30pt][r]{($\downarrow$4.90\%)}}} & 70.72\hspace{1pt}{\fontsize{7}{8}\selectfont\textcolor[gray]{0.35}{\makebox[30pt][r]{($\downarrow$5.25\%)}}} & \textbf{73.46}\hspace{1pt}{\fontsize{7}{8}\selectfont\textcolor[gray]{0.35}{\makebox[30pt][r]{($\downarrow$1.58\%)}}} & 43.2 \\
DocVQA & 91.92 & \underline{82.71}\hspace{1pt}{\fontsize{7}{8}\selectfont\textcolor[gray]{0.35}{\makebox[30pt][r]{($\downarrow$10.02\%)}}} & 78.69\hspace{1pt}{\fontsize{7}{8}\selectfont\textcolor[gray]{0.35}{\makebox[30pt][r]{($\downarrow$14.39\%)}}} & 76.57\hspace{1pt}{\fontsize{7}{8}\selectfont\textcolor[gray]{0.35}{\makebox[30pt][r]{($\downarrow$16.70\%)}}} & \textbf{91.03}\hspace{1pt}{\fontsize{7}{8}\selectfont\textcolor[gray]{0.35}{\makebox[30pt][r]{($\downarrow$0.97\%)}}} & 16.0 \\
InfoVQA & 67.44 & \underline{52.87}\hspace{1pt}{\fontsize{7}{8}\selectfont\textcolor[gray]{0.35}{\makebox[30pt][r]{($\downarrow$21.60\%)}}} & 51.75\hspace{1pt}{\fontsize{7}{8}\selectfont\textcolor[gray]{0.35}{\makebox[30pt][r]{($\downarrow$23.27\%)}}} & 50.89\hspace{1pt}{\fontsize{7}{8}\selectfont\textcolor[gray]{0.35}{\makebox[30pt][r]{($\downarrow$24.54\%)}}} & \textbf{66.11}\hspace{1pt}{\fontsize{7}{8}\selectfont\textcolor[gray]{0.35}{\makebox[30pt][r]{($\downarrow$1.97\%)}}} & 10.6 \\
ScreenSpotV2 & 89.86 & \underline{80.42}\hspace{1pt}{\fontsize{7}{8}\selectfont\textcolor[gray]{0.35}{\makebox[30pt][r]{($\downarrow$10.51\%)}}} & 72.56\hspace{1pt}{\fontsize{7}{8}\selectfont\textcolor[gray]{0.35}{\makebox[30pt][r]{($\downarrow$19.25\%)}}} & 70.44\hspace{1pt}{\fontsize{7}{8}\selectfont\textcolor[gray]{0.35}{\makebox[30pt][r]{($\downarrow$21.61\%)}}} & \textbf{89.47}\hspace{1pt}{\fontsize{7}{8}\selectfont\textcolor[gray]{0.35}{\makebox[30pt][r]{($\downarrow$0.43\%)}}} & 24.1 \\
HallusionBench & 69.88 & \underline{69.09}\hspace{1pt}{\fontsize{7}{8}\selectfont\textcolor[gray]{0.35}{\makebox[30pt][r]{($\downarrow$1.13\%)}}} & \textbf{69.26}\hspace{1pt}{\fontsize{7}{8}\selectfont\textcolor[gray]{0.35}{\makebox[30pt][r]{($\downarrow$0.89\%)}}} & 68.64\hspace{1pt}{\fontsize{7}{8}\selectfont\textcolor[gray]{0.35}{\makebox[30pt][r]{($\downarrow$1.77\%)}}} & \textbf{69.26}\hspace{1pt}{\fontsize{7}{8}\selectfont\textcolor[gray]{0.35}{\makebox[30pt][r]{($\downarrow$0.89\%)}}} & 44.0 \\
VideoOCR & 64.51 & 59.74\hspace{1pt}{\fontsize{7}{8}\selectfont\textcolor[gray]{0.35}{\makebox[30pt][r]{($\downarrow$7.39\%)}}} & \textbf{63.90}\hspace{1pt}{\fontsize{7}{8}\selectfont\textcolor[gray]{0.35}{\makebox[30pt][r]{($\downarrow$0.95\%)}}} & 63.18\hspace{1pt}{\fontsize{7}{8}\selectfont\textcolor[gray]{0.35}{\makebox[30pt][r]{($\downarrow$2.06\%)}}} & \underline{63.79}\hspace{1pt}{\fontsize{7}{8}\selectfont\textcolor[gray]{0.35}{\makebox[30pt][r]{($\downarrow$1.12\%)}}} & 47.0 \\
Video-MME & 67.78 & 67.07\hspace{1pt}{\fontsize{7}{8}\selectfont\textcolor[gray]{0.35}{\makebox[30pt][r]{($\downarrow$1.05\%)}}} & \underline{67.70}\hspace{1pt}{\fontsize{7}{8}\selectfont\textcolor[gray]{0.35}{\makebox[30pt][r]{($\downarrow$0.12\%)}}} & \textbf{67.74}\hspace{1pt}{\fontsize{7}{8}\selectfont\textcolor[gray]{0.35}{\makebox[30pt][r]{($\downarrow$0.06\%)}}} & 67.22\hspace{1pt}{\fontsize{7}{8}\selectfont\textcolor[gray]{0.35}{\makebox[30pt][r]{($\downarrow$0.83\%)}}} & 51.4 \\
LongVideoBench & 59.99 & 58.86\hspace{1pt}{\fontsize{7}{8}\selectfont\textcolor[gray]{0.35}{\makebox[30pt][r]{($\downarrow$1.88\%)}}} & \underline{60.06}\hspace{1pt}{\fontsize{7}{8}\selectfont\textcolor[gray]{0.35}{\makebox[30pt][r]{($\uparrow$0.12\%)}}} & 59.61\hspace{1pt}{\fontsize{7}{8}\selectfont\textcolor[gray]{0.35}{\makebox[30pt][r]{($\downarrow$0.63\%)}}} & \textbf{60.36}\hspace{1pt}{\fontsize{7}{8}\selectfont\textcolor[gray]{0.35}{\makebox[30pt][r]{($\uparrow$0.62\%)}}} & 49.8 \\
\midrule
\textit{Average} & 73.25 & \underline{67.87}\hspace{1pt}{\fontsize{7}{8}\selectfont\textcolor[gray]{0.35}{\makebox[30pt][r]{($\downarrow$7.11\%)}}} & 66.86\hspace{1pt}{\fontsize{7}{8}\selectfont\textcolor[gray]{0.35}{\makebox[30pt][r]{($\downarrow$7.96\%)}}} & 65.97\hspace{1pt}{\fontsize{7}{8}\selectfont\textcolor[gray]{0.35}{\makebox[30pt][r]{($\downarrow$9.08\%)}}} & \textbf{72.59}\hspace{1pt}{\fontsize{7}{8}\selectfont\textcolor[gray]{0.35}{\makebox[30pt][r]{($\downarrow$0.90\%)}}} & 35.8 \\
\midrule
\textbf{Qwen3.8-27B} & & & & & & \\
\midrule
RealWorldQA & 78.82 & 74.64\hspace{1pt}{\fontsize{7}{8}\selectfont\textcolor[gray]{0.35}{\makebox[30pt][r]{($\downarrow$5.30\%)}}} & \textbf{78.43}\hspace{1pt}{\fontsize{7}{8}\selectfont\textcolor[gray]{0.35}{\makebox[30pt][r]{($\downarrow$0.49\%)}}} & 77.65\hspace{1pt}{\fontsize{7}{8}\selectfont\textcolor[gray]{0.35}{\makebox[30pt][r]{($\downarrow$1.48\%)}}} & \underline{77.91}\hspace{1pt}{\fontsize{7}{8}\selectfont\textcolor[gray]{0.35}{\makebox[30pt][r]{($\downarrow$1.15\%)}}} & 43.1 \\
DocVQA & 92.47 & \underline{83.19}\hspace{1pt}{\fontsize{7}{8}\selectfont\textcolor[gray]{0.35}{\makebox[30pt][r]{($\downarrow$10.04\%)}}} & 77.05\hspace{1pt}{\fontsize{7}{8}\selectfont\textcolor[gray]{0.35}{\makebox[30pt][r]{($\downarrow$16.68\%)}}} & 76.01\hspace{1pt}{\fontsize{7}{8}\selectfont\textcolor[gray]{0.35}{\makebox[30pt][r]{($\downarrow$17.80\%)}}} & \textbf{91.33}\hspace{1pt}{\fontsize{7}{8}\selectfont\textcolor[gray]{0.35}{\makebox[30pt][r]{($\downarrow$1.23\%)}}} & 27.3 \\
InfoVQA & 74.13 & \underline{58.48}\hspace{1pt}{\fontsize{7}{8}\selectfont\textcolor[gray]{0.35}{\makebox[30pt][r]{($\downarrow$21.11\%)}}} & 55.49\hspace{1pt}{\fontsize{7}{8}\selectfont\textcolor[gray]{0.35}{\makebox[30pt][r]{($\downarrow$25.15\%)}}} & 53.96\hspace{1pt}{\fontsize{7}{8}\selectfont\textcolor[gray]{0.35}{\makebox[30pt][r]{($\downarrow$27.21\%)}}} & \textbf{71.61}\hspace{1pt}{\fontsize{7}{8}\selectfont\textcolor[gray]{0.35}{\makebox[30pt][r]{($\downarrow$3.40\%)}}} & 16.5 \\
ScreenSpotV2 & 94.42 & \underline{91.04}\hspace{1pt}{\fontsize{7}{8}\selectfont\textcolor[gray]{0.35}{\makebox[30pt][r]{($\downarrow$3.58\%)}}} & 49.53\hspace{1pt}{\fontsize{7}{8}\selectfont\textcolor[gray]{0.35}{\makebox[30pt][r]{($\downarrow$47.54\%)}}} & 49.21\hspace{1pt}{\fontsize{7}{8}\selectfont\textcolor[gray]{0.35}{\makebox[30pt][r]{($\downarrow$47.88\%)}}} & \textbf{91.43}\hspace{1pt}{\fontsize{7}{8}\selectfont\textcolor[gray]{0.35}{\makebox[30pt][r]{($\downarrow$3.17\%)}}} & 55.1 \\
HallusionBench & 73.25 & \underline{72.10}\hspace{1pt}{\fontsize{7}{8}\selectfont\textcolor[gray]{0.35}{\makebox[30pt][r]{($\downarrow$1.57\%)}}} & 71.92\hspace{1pt}{\fontsize{7}{8}\selectfont\textcolor[gray]{0.35}{\makebox[30pt][r]{($\downarrow$1.82\%)}}} & 71.66\hspace{1pt}{\fontsize{7}{8}\selectfont\textcolor[gray]{0.35}{\makebox[30pt][r]{($\downarrow$2.17\%)}}} & \textbf{73.60}\hspace{1pt}{\fontsize{7}{8}\selectfont\textcolor[gray]{0.35}{\makebox[30pt][r]{($\uparrow$0.48\%)}}} & 44.9 \\
VideoOCR & 70.72 & 65.59\hspace{1pt}{\fontsize{7}{8}\selectfont\textcolor[gray]{0.35}{\makebox[30pt][r]{($\downarrow$7.25\%)}}} & \underline{69.69}\hspace{1pt}{\fontsize{7}{8}\selectfont\textcolor[gray]{0.35}{\makebox[30pt][r]{($\downarrow$1.46\%)}}} & 69.54\hspace{1pt}{\fontsize{7}{8}\selectfont\textcolor[gray]{0.35}{\makebox[30pt][r]{($\downarrow$1.67\%)}}} & \textbf{70.51}\hspace{1pt}{\fontsize{7}{8}\selectfont\textcolor[gray]{0.35}{\makebox[30pt][r]{($\downarrow$0.30\%)}}} & 49.6 \\
Video-MME & 70.89 & 70.41\hspace{1pt}{\fontsize{7}{8}\selectfont\textcolor[gray]{0.35}{\makebox[30pt][r]{($\downarrow$0.68\%)}}} & 70.70\hspace{1pt}{\fontsize{7}{8}\selectfont\textcolor[gray]{0.35}{\makebox[30pt][r]{($\downarrow$0.27\%)}}} & \underline{70.96}\hspace{1pt}{\fontsize{7}{8}\selectfont\textcolor[gray]{0.35}{\makebox[30pt][r]{($\uparrow$0.10\%)}}} & \textbf{71.04}\hspace{1pt}{\fontsize{7}{8}\selectfont\textcolor[gray]{0.35}{\makebox[30pt][r]{($\uparrow$0.21\%)}}} & 54.2 \\
LongVideoBench & 66.49 & 64.70\hspace{1pt}{\fontsize{7}{8}\selectfont\textcolor[gray]{0.35}{\makebox[30pt][r]{($\downarrow$2.69\%)}}} & \underline{65.74}\hspace{1pt}{\fontsize{7}{8}\selectfont\textcolor[gray]{0.35}{\makebox[30pt][r]{($\downarrow$1.13\%)}}} & 65.45\hspace{1pt}{\fontsize{7}{8}\selectfont\textcolor[gray]{0.35}{\makebox[30pt][r]{($\downarrow$1.56\%)}}} & \textbf{66.42}\hspace{1pt}{\fontsize{7}{8}\selectfont\textcolor[gray]{0.35}{\makebox[30pt][r]{($\downarrow$0.11\%)}}} & 53.0 \\
\midrule
\textit{Average} & 77.65 & \underline{72.52}\hspace{1pt}{\fontsize{7}{8}\selectfont\textcolor[gray]{0.35}{\makebox[30pt][r]{($\downarrow$6.53\%)}}} & 67.32\hspace{1pt}{\fontsize{7}{8}\selectfont\textcolor[gray]{0.35}{\makebox[30pt][r]{($\downarrow$11.82\%)}}} & 66.81\hspace{1pt}{\fontsize{7}{8}\selectfont\textcolor[gray]{0.35}{\makebox[30pt][r]{($\downarrow$12.46\%)}}} & \textbf{76.73}\hspace{1pt}{\fontsize{7}{8}\selectfont\textcolor[gray]{0.35}{\makebox[30pt][r]{($\downarrow$1.08\%)}}} & 43.0 \\
\bottomrule
\end{tabular*}
\end{table}

  With a default routing threshold $t=0.5$, VisionWeave achieves
  content-adaptive token savings while maintaining near-native quality
  (Table~\ref{tab:exp-main}). 
  Across all eight benchmarks, VisionWeave incurs an average performance
  loss of only 0.90\% on Qwen3.5-4B and 1.08\% on Qwen3.8-27B,
  relative to the corresponding native models. 
  Specifically, on Qwen3.8-27B, VisionWeave saves 16.5\% of tokens on InfoVQA
  and 27.3\% on DocVQA, compared with 49.6--54.2\% across the
  three video benchmarks, where performance losses remain at or
  below 0.30\%. Qwen3.5-4B shows a similar pattern, saving
  10.6--16.0\% on the two document benchmarks and 47.0--51.4\%
  on video.

In contrast, FastV$^\dagger$ and VisionZip incur average
  losses of 11.82\% and 12.46\% on Qwen3.8-27B at a fixed
  50\% token-savings target. Although losses are modest on
  video, they degrade sharply on grounding and information-dense 
  tasks: VisionZip incurs losses of 47.88\% on ScreenSpotV2 and
  17.80\% on DocVQA. VisionWeave, by contrast, loses only 3.17\%
  on ScreenSpotV2 despite more aggressive token savings of 55.1\%,
  while adaptively reducing its savings to 27.3\% on DocVQA with
  only a 1.23\% loss. These results highlight the advantage of
  content-adaptive savings: reducing visual tokens according to
  input demands while preserving quality.

  To isolate the effect of dataset-level savings, we also assign
  FastV$^\dagger$ and VisionZip \method{}'s observed savings ratio for
  each backbone and dataset. \method{} retains the highest average
  score: 72.59 versus 72.20/71.90 on Qwen3.5-4B, and 76.73 versus
  70.10/70.33 on Qwen3.8-27B. Rankings vary across tasks, but matching
  the dataset-level savings alone does not reproduce \method{}'s
  average quality. Complete results appear in Appendix~\ref{sec:exp-matched}.

\begin{figure}[!t]
\centering
\vspace{-0.2cm}
\includegraphics[width=\linewidth]{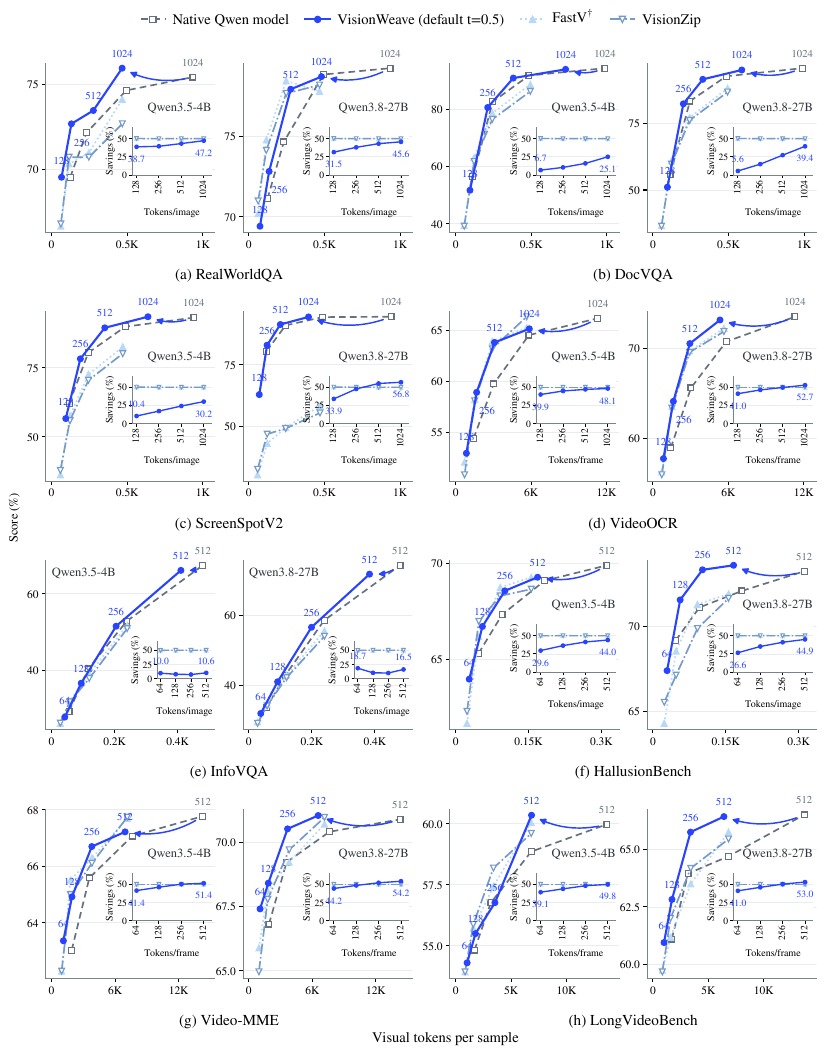}\par\nointerlineskip
\vspace{-0.2cm}
\caption{\method{} achieves consistently favorable trade-offs across tasks and input resolutions.}
\label{fig:exp-resolution-robustness}
\end{figure}

\subsection{Robustness Across Input Resolutions and Frame Budgets}
\label{sec:exp-resolution}

\begin{figure}[!t]
\centering
\begin{minipage}[t]{0.55\linewidth}
\vspace{-0.2cm}
\centering
\includegraphics[width=\linewidth]{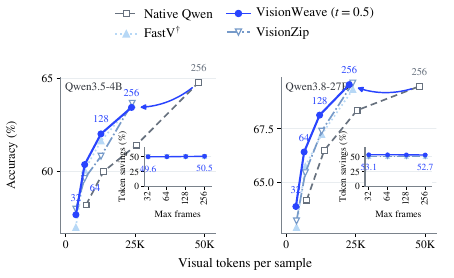}
\caption{\method{} maintains favorable trade-offs across frame budgets on LongVideoBench.}
\vspace{-0.2cm}
\label{fig:exp-frame-robustness}
\end{minipage}\hfill
\begin{minipage}[t]{0.43\linewidth}
\vspace{0pt}
\centering
\includegraphics[width=\linewidth]{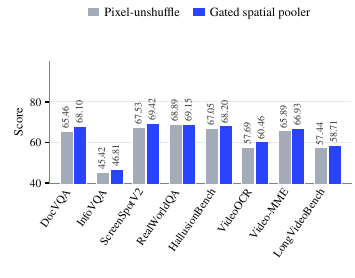}
\caption{Gated pooling outperforms pixel-unshuffle across eight benchmarks.}
\vspace{-0.2cm}

\label{fig:exp-pooler-ablation}
\end{minipage}
\end{figure}

We next demonstrate the robustness of \method{}'s efficiency--quality
  trade-offs across tasks, spatial resolutions, and frame budgets.
  Consistently improving upon input downsampling is nontrivial:
  a method that works well in one setting may fail in another.
  Figure~\ref{fig:exp-resolution-robustness} shows that \method{} consistently
  maintains favorable trade-offs across all eight benchmarks and
  the evaluated resolutions. Video resolution sweeps fix the frame cap
  at 64 and the sampling rate at 2 FPS. FastV$^\dagger$ and VisionZip remain competitive
  on VideoOCR yet degrade sharply on ScreenSpotV2.
  Their effectiveness also varies with resolution within the same task.
  On Qwen3.8-27B DocVQA, for example, VisionZip offers a competitive
  trade-off at a 256-token input budget but falls behind simple
  downsampling at higher resolutions.
  By contrast,  \method{} consistently offers a more favorable
  efficiency--quality trade-off than input downsampling across
  the evaluated tasks and resolutions.
  The same advantage extends to temporal coverage
  (Figure~\ref{fig:exp-frame-robustness}). With a 256-frame cap,
  Qwen3.8-27B \method{} outperforms the native model capped at
  128 frames by 1.20 points on LongVideoBench while using fewer
  visual tokens.
  Together, these results support the robust performance of native elastic visual weaving
  across diverse input conditions.

\subsection{End-to-End Serving Efficiency on SGLang}
\label{sec:exp-serving}

\begin{table}[!t]
\vspace{-0.5cm}
\captionof{table}[!t]{\method{} accelerates Qwen3.8-27B serving on SGLang.}
\label{tab:exp-serving}
\centering
\fontsize{8}{9.6}\selectfont
\setlength{\tabcolsep}{2.0pt}
\renewcommand{\arraystretch}{1.25}
\begin{tabular*}{\linewidth}{@{\extracolsep{\fill}}lccccc@{}}
\toprule
Model & Throughput (req/min) $\uparrow$ & Mean TTFT (s) $\downarrow$ & P95 TTFT (s) $\downarrow$ & Mean TPOT (ms) $\downarrow$ & P95 TPOT (ms) $\downarrow$ \\
\midrule
Native Model& 0.90 & 169.90 & 281.41 & 355.64 & 715.58 \\
\method{} & \textbf{2.07}\hspace{1pt}{\fontsize{6}{7.2}\selectfont\textcolor[gray]{0.35}{\makebox[28pt][r]{($2.30\times$)}}}
& \textbf{77.54}\hspace{1pt}{\fontsize{6}{7.2}\selectfont\textcolor[gray]{0.35}{\makebox[28pt][r]{($\downarrow$54.4\%)}}}
& \textbf{120.59}\hspace{1pt}{\fontsize{6}{7.2}\selectfont\textcolor[gray]{0.35}{\makebox[28pt][r]{($\downarrow$57.1\%)}}}
& \textbf{140.29}\hspace{1pt}{\fontsize{6}{7.2}\selectfont\textcolor[gray]{0.35}{\makebox[28pt][r]{($\downarrow$60.6\%)}}}
& \textbf{289.19}\hspace{1pt}{\fontsize{6}{7.2}\selectfont\textcolor[gray]{0.35}{\makebox[28pt][r]{($\downarrow$59.6\%)}}} \\
\bottomrule
\end{tabular*}

\end{table}

To evaluate practical deployability, we benchmark native Qwen3.8-27B
and \method{} on SGLang using our integration
(Section~\ref{sec:serving}). Both run on two A100-80GB GPUs
(BF16, TP=2), with chunked prefill and CUDA-graph decoding enabled.
We use the  LongVideoBench with fps2, 256 max frames per video,
a spatial budget of 1024 tokens/frame. Client concurrency is four, with a server limit of eight.
More details appear
in Appendix~\ref{sec:app-serving}. Overall, \method{} delivers
a $2.30\times$ end-to-end throughput speedup
(Table~\ref{tab:exp-serving}). Both mean and tail latency improve:
P95 time to first token (TTFT) and time per output token (TPOT)
decrease by 57.1\% and 59.6\%, respectively.
These results demonstrate that native elastic visual processing
translates into higher serving throughput and lower response latency
under this long-video workload.

\subsection{Inference-Time Control of Token Savings}
\label{sec:exp-control}

\begin{figure}[!t]
\centering
\includegraphics[width=\linewidth]{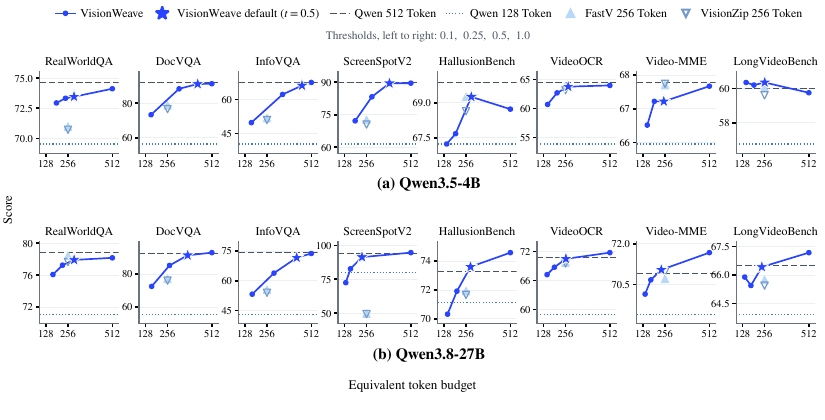}
\vspace{-1.0cm}
\caption{VisionWeave enables flexible inference-time control of token savings while closely matching native-model performance when savings are disabled.}
\label{fig:exp-thresholds}
\end{figure}

 The default threshold $t=0.5$ already provides content-adaptive
  token savings without task-specific tuning. We further show that
  varying this threshold enables flexible inference-time control
  without retraining (Figure~\ref{fig:exp-thresholds}).
  At the full-retention endpoint, labeled $t=1$ in the figure,
  every block uses its fine-grained representation, with no token
  savings. 
  These results suggest that native visual weaving expands the
  model's efficiency--quality operating range while 
  preserving its fully fine-grained performance. Lowering the threshold then favors coarse-grained representations
  and increases token savings. On Video-MME, reducing $t$ from
  $0.5$ to $0.25$ increases savings from 54.2\% to 66.0\% on
  Qwen3.8-27B, with only a 0.37-point performance loss.
  On Qwen3.5-4B, the same adjustment increases savings from
  51.4\% to 62.1\% while preserving the score of 67.22.
  These results highlight a key practical advantage of native visual
  weaving: a single model can accommodate different quality and cost
  requirements without retraining, while closely matching the native
  model's performance ceiling when token savings are disabled.

\FloatBarrier
\raggedbottom
\section{Analysis of Elastic Visual Weaving}
\label{sec:analysis}

We examine the contributions of individual components and the
spatial allocation of visual representations.

\subsection{Ablation Studies}
\label{sec:analysis-ablation}
\label{sec:exp-pooler-ablation}
\label{sec:exp-router-ablation}

\paragraph{Gated spatial pooler.}
We compare our pooler with a pixel-unshuffle-based projector on
Qwen3.5-4B after Stage~1, both at 75\% token savings.
Our pooler achieves higher scores on all eight benchmarks
(Figure~\ref{fig:exp-pooler-ablation}), including gains of 2.64 points
on DocVQA and 2.77 points on VideoOCR. This supports gated spatial
pooling for constructing coarse-grained visual representations.

\paragraph{Learned granularity allocation.}
Using the main comparison's $512$-token image/frame input budgets,
we shuffle block-wise routing scores within each image or video before
applying $t=0.5$. This preserves the number of coarse selections for
each input, including input-dependent token savings, but breaks their
alignment with visual content. Mean scores decrease by $2.31$ and
$5.36$ points on Qwen3.5-4B and Qwen3.8-27B, respectively
(Table~\ref{tab:app-router-ablation}). Even excluding ScreenSpotV2,
the mean decreases remain $1.30$ and $1.53$ points. Thus, learning
where to allocate each granularity matters beyond choosing how many
tokens to retain.

\begin{table}[!htbp]
\caption{Learned granularity allocation improves performance over shuffled routing.}
\label{tab:app-router-ablation}
\centering
\small
\setlength{\tabcolsep}{4pt}
\renewcommand{\arraystretch}{1.05}
\begin{tabular*}{\linewidth}{@{\extracolsep{\fill}}lrrrrrr@{}}
\toprule
& \multicolumn{3}{c}{\textbf{Qwen3.5-4B}} & \multicolumn{3}{c}{\textbf{Qwen3.8-27B}} \\
\cmidrule(lr){2-4}\cmidrule(l){5-7}
Benchmark & VisionWeave & Shuffled & $\Delta$ & VisionWeave & Shuffled & $\Delta$ \\
\midrule
RealWorldQA & \textbf{73.46} & \textbf{73.46} & $0.00$ & \textbf{77.91} & 76.60 & $-1.31$ \\
DocVQA & \textbf{91.03} & 88.70 & $-2.32$ & \textbf{91.33} & 88.35 & $-2.98$ \\
InfoVQA & \textbf{66.11} & 64.83 & $-1.28$ & \textbf{71.61} & 69.04 & $-2.58$ \\
ScreenSpotV2 & \textbf{89.47} & 80.03 & $-9.43$ & \textbf{91.43} & 59.28 & $-32.15$ \\
HallusionBench & \textbf{69.26} & 67.32 & $-1.95$ & \textbf{73.60} & 71.66 & $-1.95$ \\
VideoOCR & \textbf{63.79} & 61.85 & $-1.95$ & \textbf{70.51} & 69.44 & $-1.08$ \\
Video-MME & \textbf{67.22} & 66.78 & $-0.44$ & \textbf{71.04} & 70.67 & $-0.37$ \\
LongVideoBench & \textbf{60.36} & 59.24 & $-1.12$ & \textbf{66.42} & 65.97 & $-0.45$ \\
\midrule
\textit{Average} & \textbf{72.59} & 70.28 & $-2.31$ & \textbf{76.73} & 71.37 & $-5.36$ \\
\bottomrule
\end{tabular*}
\smallskip
\begin{minipage}{\linewidth}\footnotesize
\end{minipage}
\end{table}

\subsection{Visual Analysis of Granularity Allocation}
\label{sec:analysis-visual}

\begin{figure}[!t]
\centering
\includegraphics[width=\linewidth]{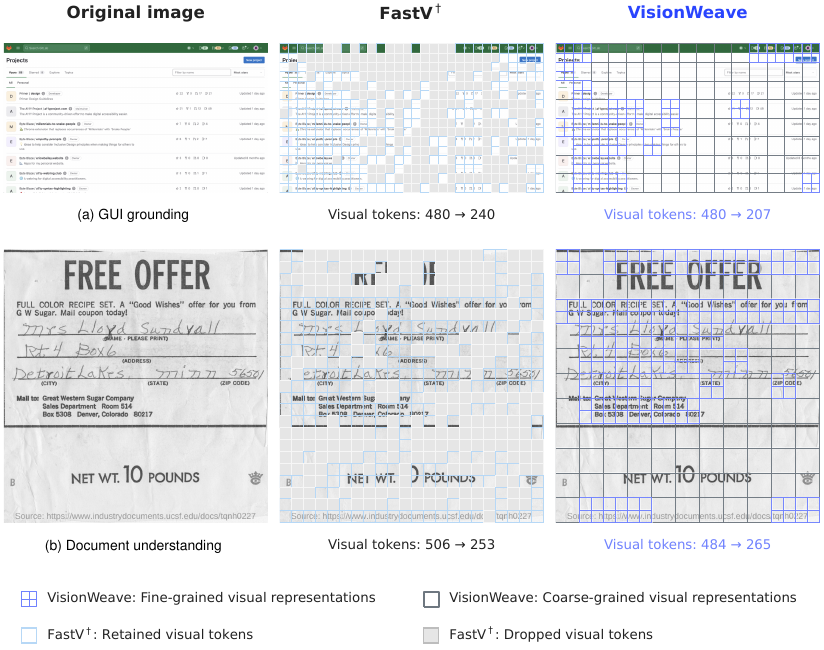}
\caption{\method{} maintains spatial coverage through mixed granularity, whereas FastV$^\dagger$ drops tokens covering task-relevant content.}
\label{fig:granularity-case-studies}
\end{figure}

Figure~\ref{fig:granularity-case-studies} compares \method{} and
FastV$^\dagger$ on GUI grounding and DocVQA examples using Qwen3.8-27B
at a 512-token input budget. In the web interface, FastV$^\dagger$
drops tokens covering click targets, including the star button,
whereas \method{} keeps every region represented. In the document,
\method{} saves $45.2\%$ of tokens while representing larger text
coarsely; FastV$^\dagger$ removes tokens covering informative text,
leaving some characters only partially represented. These examples
illustrate how changing granularity maintains spatial coverage
instead of leaving gaps, helping explain the robustness on grounding
and information-dense inputs. Additional cases appear in
Appendix~\ref{sec:app-case-studies}.

\section{Conclusion }

  We introduced \method{}, establishing elastic visual representation
  weaving as a native capability of MLLMs through self-distillation alone.
  It combines content-adaptive token savings with robust
  efficiency--quality trade-offs across tasks, resolutions, and frame budgets.
  Our SGLang integration further translates these token savings into higher
  throughput and lower latency on modern serving infrastructure.
  These findings establish representational granularity as a learnable degree of freedom, validating content-adaptive computational effort as a feasible architectural principle.

\section{Acknowledge}
This work was supported by Alibaba Research Intern Program. We would like to thank the Qwen
Team at Alibaba Token Hub (ATH), Alibaba Group, for providing the computational resources and
foundation models (Qwen) used in this research.

\bibliography{refs}

@article{Yang2024VisionZipLI,
  title={VisionZip: Longer is Better but Not Necessary in Vision Language Models},
  author={Senqiao Yang and Yukang Chen and Zhuotao Tian and Chengyao Wang and Jingyao Li and Bei Yu and Jiaya Jia},
  journal={2025 IEEE/CVF Conference on Computer Vision and Pattern Recognition (CVPR)},
  year={2024},
  pages={19792-19802},
  url={https://api.semanticscholar.org/CorpusID:274514545}
}

@inproceedings{Chen2024AnII,
  title={An Image is Worth 1/2 Tokens After Layer 2: Plug-and-Play Inference Acceleration for Large Vision-Language Models},
  author={Liang Chen and Haozhe Zhao and Tianyu Liu and Shuai Bai and Junyang Lin and Chang Zhou and Baobao Chang},
  booktitle={European Conference on Computer Vision},
  year={2024},
  url={https://api.semanticscholar.org/CorpusID:268358224}
}

@article{Ye2024ATPLLaVAAT,
  title={ATP-LLaVA: Adaptive Token Pruning for Large Vision Language Models},
  author={Xubing Ye and Yukang Gan and Yi-Xiao Ge and Xiao-Ping Zhang and Yansong Tang},
  journal={2025 IEEE/CVF Conference on Computer Vision and Pattern Recognition (CVPR)},
  year={2024},
  pages={24972-24982},
  url={https://api.semanticscholar.org/CorpusID:274436316}
}

@article{Cui2025ViCOAT,
  title={ViCO: A Training Strategy towards Semantic Aware Dynamic High-Resolution},
  author={Long Cui and Weiyun Wang and Jie Shao and Zichen Wen and Gen Luo and Linfeng Zhang and Yanting Zhang and Yu Qiao and Wenhai Wang},
  journal={ArXiv},
  year={2025},
  volume={abs/2510.12793},
  url={https://api.semanticscholar.org/CorpusID:282064724}
}

@article{Bai2025Qwen25VLTR,
  title={Qwen2.5-VL Technical Report},
  author={Shuai Bai and Ke-qin Chen and Xue-Jing Liu and Jia-Lin Wang and Wenbin Ge and Sibo Song and Kai Dang and Peng Wang and Shijie Wang and Jun Tang and Humen Zhong and Yuanzhi Zhu and Ming-Hsuan Yang and Zhaohai Li and Jian-Qiang Wan and Peng-Fei Wang and Wei Ding and Zheren Fu and Yiheng Xu and Jiabo Ye and Xi Zhang and Tianbao Xie and Zesen Cheng and Hang Zhang and Zhibo Yang and Haiyang Xu and Junyang Lin},
  journal={ArXiv},
  year={2025},
  volume={abs/2502.13923},
  url={https://api.semanticscholar.org/CorpusID:276449796}
}

@article{Bai2025Qwen3VLTR,
  title={Qwen3-VL Technical Report},
  author={Shuai Bai and Yuxuan Cai and Rui-Zhe Chen and Ke-qin Chen and Xiong-Hui Chen and Zesen Cheng and Liang-Hao Deng and Wei Ding and Rongyao Fang and Chang Gao and Chunjiang Ge and Wenbin Ge and Zhifang Guo and Qidong Huang and Qidong Huang and Fei Huang and Binyuan Hui and Shutong Jiang and Zhaohai Li and Ming-Sheng Li and Mei Li and Kaixin Li and Zicheng Lin and Junyang Lin and Xue-Jing Liu and Jiawei Liu and Cheng-Long Liu and Yang Liu and Dayiheng Liu and Shixuan Liu and Dunjie Lu and Ruilin Luo and Chenxu Lv and Rui Men and Lingchen Meng and Lingchen Meng and Xin-yi Ren and Sibo Song and Yu-chen Sun and Jun Tang and Jianhong Tu and Jian-Qiang Wan and Peng Wang and Pengfei Wang and Qiu-Yue Wang and Yuxuan Wang and Tianbao Xie and Yihe Xu and Haiyang Xu and Jin Xu and Zhibo Yang and Mingkun Yang and Jianxin Yang and An Yang and Bo-Wen Yu and Fei Zhang and Hang Zhang and Xi Zhang and Botao Zheng and Humen Zhong and Jingren Zhou and Fanxi Zhou and Jingren Zhou and Yuanzhi Zhu and Keming Zhu},
  journal={ArXiv},
  year={2025},
  volume={abs/2511.21631},
  url={https://api.semanticscholar.org/CorpusID:283262018}
}

@article{Shao2025WhenTT,
  title={When Tokens Talk Too Much: A Survey of Multimodal Long-Context Token Compression across Images, Videos, and Audios},
  author={Kele Shao and Keda Tao and Kejia Zhang and Sicheng Feng and Mu Cai and Yuzhang Shang and Haoxuan You and Can Qin and Yang Sui and Huan Wang},
  journal={ArXiv},
  year={2025},
  volume={abs/2507.20198},
  url={https://api.semanticscholar.org/CorpusID:290088866}
}

@article{Tao2024DyCokeD,
  title={DyCoke : Dynamic Compression of Tokens for Fast Video Large Language Models},
  author={Keda Tao and Can Qin and Haoxuan You and Yang Sui and Huan Wang},
  journal={2025 IEEE/CVF Conference on Computer Vision and Pattern Recognition (CVPR)},
  year={2024},
  pages={18992-19001},
  url={https://api.semanticscholar.org/CorpusID:274192345}
}

@article{Shao2025ASO,
  title={A Survey of Token Compression for Efficient Multimodal Large Language Models},
  author={Kele Shao and Keda Tao and Kejia Zhang and Sicheng Feng and Mu Cai and Yuzhang Shang and Haoxuan You and Can Qin and Yang Sui and Huan Wang},
  journal={Trans. Mach. Learn. Res.},
  year={2025},
  volume={2026},
  url={https://api.semanticscholar.org/CorpusID:280323457}
}

@article{Yu2025DynamicGM,
  title={Dynamic Granularity Matters: Rethinking Vision Transformers Beyond Fixed Patch Splitting},
  author={Qiyang Yu and Yu Fang and Tianrui Li and Xuemei Cao and Yan Chen and Jianghao Li and Fan Min},
  journal={ArXiv},
  year={2025},
  volume={abs/2511.19021},
  url={https://api.semanticscholar.org/CorpusID:283244585}
}

@article{Choudhury2025AcceleratingVT,
  title={Accelerating Vision Transformers with Adaptive Patch Sizes},
  author={Rohan Choudhury and Jungeun Kim and Jinhyung Park and Eunho Yang and L{\'a}szl{\'o} A. Jeni and Kris M. Kitani},
  journal={ArXiv},
  year={2025},
  volume={abs/2510.18091},
  url={https://api.semanticscholar.org/CorpusID:282246197}
}

@inproceedings{Bai2026KimiKO,
  title={Kimi K3: Open Frontier Intelligence},
  author={Kimi Team},
  year={2026},
  url={https://api.semanticscholar.org/CorpusID:290625162}
}

@inproceedings{DeepSeekAI2026DeepSeekV4TH,
  title={DeepSeek-V4: Towards Highly Efficient Million-Token Context Intelligence},
  author={DeepSeek-AI },
  year={2026},
  url={https://api.semanticscholar.org/CorpusID:289623513}
}

@article{Zheng2023SGLangEE,
  title={SGLang: Efficient Execution of Structured Language Model Programs},
  author={Lianmin Zheng and Liangsheng Yin and Zhiqiang Xie and Chu-Yue Sun and Jeff Huang and Cody Hao Yu and Shiyi Cao and Christoforos E. Kozyrakis and Ion Stoica and Joseph E. Gonzalez and Clark W. Barrett and Ying Sheng},
  journal={Advances in Neural Information Processing Systems 37},
  year={2023},
  url={https://api.semanticscholar.org/CorpusID:266174771}
}

@article{Peng2023YaRNEC,
  title={YaRN: Efficient Context Window Extension of Large Language Models},
  author={Bowen Peng and Jeffrey Quesnelle and Honglu Fan and Enrico Shippole},
  journal={ArXiv},
  year={2023},
  volume={abs/2309.00071},
  url={https://api.semanticscholar.org/CorpusID:261493986}
}

@article{Chen2023ExtendingCW,
  title={Extending Context Window of Large Language Models via Positional Interpolation},
  author={Shouyuan Chen and Sherman Wong and Liangjian Chen and Yuandong Tian},
  journal={ArXiv},
  year={2023},
  volume={abs/2306.15595},
  url={https://api.semanticscholar.org/CorpusID:259262376}
}

@misc{an2025llavaonevision15fullyopenframework,
      title={LLaVA-OneVision-1.5: Fully Open Framework for Democratized Multimodal Training}, 
      author={Xiang An and Yin Xie and Kaicheng Yang and Wenkang Zhang and Xiuwei Zhao and Zheng Cheng and Yirui Wang and Songcen Xu and Changrui Chen and Didi Zhu and Chunsheng Wu and Huajie Tan and Chunyuan Li and Jing Yang and Jie Yu and Xiyao Wang and Bin Qin and Yumeng Wang and Zizhen Yan and Ziyong Feng and Ziwei Liu and Bo Li and Jiankang Deng},
      year={2025},
      eprint={2509.23661},
      archivePrefix={arXiv},
      primaryClass={cs.CV},
      url={https://arxiv.org/abs/2509.23661}, 
}

@misc{zhang2025llavavideovideoinstructiontuning,
      title={LLaVA-Video: Video Instruction Tuning With Synthetic Data}, 
      author={Yuanhan Zhang and Jinming Wu and Wei Li and Bo Li and Zejun Ma and Ziwei Liu and Chunyuan Li},
      year={2025},
      eprint={2410.02713},
      archivePrefix={arXiv},
      primaryClass={cs.CV},
      url={https://arxiv.org/abs/2410.02713}, 
}

@misc{gou2025navigatingdigitalworldhumans,
      title={Navigating the Digital World as Humans Do: Universal Visual Grounding for GUI Agents}, 
      author={Boyu Gou and Ruohan Wang and Boyuan Zheng and Yanan Xie and Cheng Chang and Yiheng Shu and Huan Sun and Yu Su},
      year={2025},
      eprint={2410.05243},
      archivePrefix={arXiv},
      primaryClass={cs.AI},
      url={https://arxiv.org/abs/2410.05243}, 
}

@misc{guo2026videommadaptivemacromicroinference,
      title={VideoMM: Adaptive Macro-Micro Inference for Efficient Video MLLMs}, 
      author={Haoyu Guo and Yuan Feng and Junlin Lv and Mingjun Xiao and S Kevin Zhou and Xike Xie},
      year={2026},
      eprint={2609.16722},
      archivePrefix={arXiv},
      primaryClass={cs.AI},
      url={https://arxiv.org/abs/2609.16722}, 
}

@misc{chen2026variationawarevisiontokendropping,
      title={Variation-aware Vision Token Dropping for Faster Large Vision-Language Models}, 
      author={Junjie Chen and Xuyang Liu and Zichen Wen and Yiyu Wang and Siteng Huang and Honggang Chen},
      year={2026},
      eprint={2509.01552},
      archivePrefix={arXiv},
      primaryClass={cs.CV},
      url={https://arxiv.org/abs/2509.01552}, 
}

@misc{han2025filtercorrelatecompresstrainingfree,
      title={Filter, Correlate, Compress: Training-Free Token Reduction for MLLM Acceleration}, 
      author={Yuhang Han and Xuyang Liu and Zihan Zhang and Pengxiang Ding and Junjie Chen and Donglin Wang and Honggang Chen and Qingsen Yan and Siteng Huang},
      year={2025},
      eprint={2411.17686},
      archivePrefix={arXiv},
      primaryClass={cs.CV},
      url={https://arxiv.org/abs/2411.17686}, 
}

@article{Jeddi2025SimilarityAwareTP,
  title={Similarity-Aware Token Pruning: Your VLM but Faster},
  author={Ahmadreza Jeddi and Negin Baghbanzadeh and Elham Dolatabadi and Babak Taati},
  journal={ArXiv},
  year={2025},
  volume={abs/2503.11549},
  url={https://api.semanticscholar.org/CorpusID:277043961}
}

@article{Zhang2025BeyondAO,
  title={Beyond Attention or Similarity: Maximizing Conditional Diversity for Token Pruning in MLLMs},
  author={Qizhe Zhang and Meng-Zhen Liu and Li-Chen Li and Ming Lu and Yuan Zhang and Junwen Pan and Qi She and Shanghang Zhang},
  journal={ArXiv},
  year={2025},
  volume={abs/2506.10967},
  url={https://api.semanticscholar.org/CorpusID:279318547}
}

@misc{cai2024matryoshkamultimodalmodels,
      title={Matryoshka Multimodal Models}, 
      author={Mu Cai and Jianwei Yang and Jianfeng Gao and Yong Jae Lee},
      year={2024},
      eprint={2405.17430},
      archivePrefix={arXiv},
      primaryClass={cs.CV},
      url={https://arxiv.org/abs/2405.17430}, 
}

@misc{hu2024matryoshkaquerytransformerlarge,
      title={Matryoshka Query Transformer for Large Vision-Language Models}, 
      author={Wenbo Hu and Zi-Yi Dou and Liunian Harold Li and Amita Kamath and Nanyun Peng and Kai-Wei Chang},
      year={2024},
      eprint={2405.19315},
      archivePrefix={arXiv},
      primaryClass={cs.CV},
      url={https://arxiv.org/abs/2405.19315}, 
}

@misc{kuzucu2026parcelpoolanchoredresamplingconditioned,
      title={PARCEL: Pool-Anchored Resampling with Conditioned Elastic Queries for Efficient Vision-Language Understanding}, 
      author={Selim Kuzucu and Alessio Tonioni and Vasile Lup and Bernt Schiele and Federico Tombari and Muhammad Ferjad Naeem},
      year={2026},
      eprint={2605.30126},
      archivePrefix={arXiv},
      primaryClass={cs.CV},
      url={https://arxiv.org/abs/2605.30126}, 
}

@article{Feng2024AdaKVOK,
  title={Ada-KV: Optimizing KV Cache Eviction by Adaptive Budget Allocation for Efficient LLM Inference},
  author={Yuan Feng and Junlin Lv and Yukun Cao and Xike Xie and S. Kevin Zhou},
  journal={ArXiv},
  year={2024},
  volume={abs/2407.11550},
  url={https://api.semanticscholar.org/CorpusID:271218006}
}

@inproceedings{Feng2025CriticalKVOK,
  title={CriticalKV: Optimizing KV Cache Eviction from an Output Perturbation Perspective},
  author={Yuan Feng and Junlin Lv and Yukun Cao and Xike Xie and S. Kevin Zhou},
  year={2025},
  url={https://api.semanticscholar.org/CorpusID:276161406}
}

@article{Wang2025SparseMMHS,
  title={SparseMM: Head Sparsity Emerges from Visual Concept Responses in MLLMs},
  author={Jiahui Wang and Zuyan Liu and Yongming Rao and Jiwen Lu},
  journal={2025 IEEE/CVF International Conference on Computer Vision (ICCV)},
  year={2025},
  pages={23177-23187},
  url={https://api.semanticscholar.org/CorpusID:279244559}
}

@article{Liu2025MixingIW,
  title={Mixing Importance with Diversity: Joint Optimization for KV Cache Compression in Large Vision-Language Models},
  author={Xu-Yang Liu and Xiyan Gui and Yuchao Zhang and Linfeng Zhang},
  journal={ArXiv},
  year={2025},
  volume={abs/2510.20707},
  url={https://api.semanticscholar.org/CorpusID:282304428}
}

@article{Feng2025TamingTF,
  title={Taming the Fragility of KV Cache Eviction in LLM Inference},
  author={Yuan Feng and Hao-Yu Guo and Junlin Lv and S. Kevin Zhou and Xike Xie},
  journal={ArXiv},
  year={2025},
  volume={abs/2510.13334},
  url={https://api.semanticscholar.org/CorpusID:282102254}
}
\bibliographystyle{iclr2026_conference}

\clearpage
\appendix
\etocdepthtag.toc{appendix}
\begingroup
\etocsettagdepth{main}{none}
\etocsettagdepth{appendix}{subsection}
\etocsettocstyle{\section*{Appendix Contents}}{}
\tableofcontents
\endgroup
\clearpage

\section{Limitations and Future Works}

Our ablations examine on- versus off-policy self-distillation, the
  contribution of Stage~3, and the pooler architecture. However, the
  substantial computational cost of training limits a more comprehensive
  exploration of architectural choices, including alternative router
  designs. These unexplored choices offer opportunities to further improve
  elastic visual weaving. This work focuses on two granularities, while we do not foresee significant  barriers to extending our approach to diverse levels. We therefore expect that moving beyond the current two-level design to richer multi-scale representations could yield further gains. Moreover, whereas we currently implant this capability via post-training self-distillation, future work could explore introducing elastic visual weaving during base model pre-training. Our long-term vision is a new generation of multimodal foundation models that jointly develop visual understanding and efficiency, natively modulating computational effort by information content.

\section{Case Studies}
\label{sec:app-case-studies}
\subsection{Case Study in GUI Grounding Tasks}
\label{sec:app-gui-visualizations}

\begin{figure}[!ht]
  \centering
  \includegraphics[width=\linewidth]{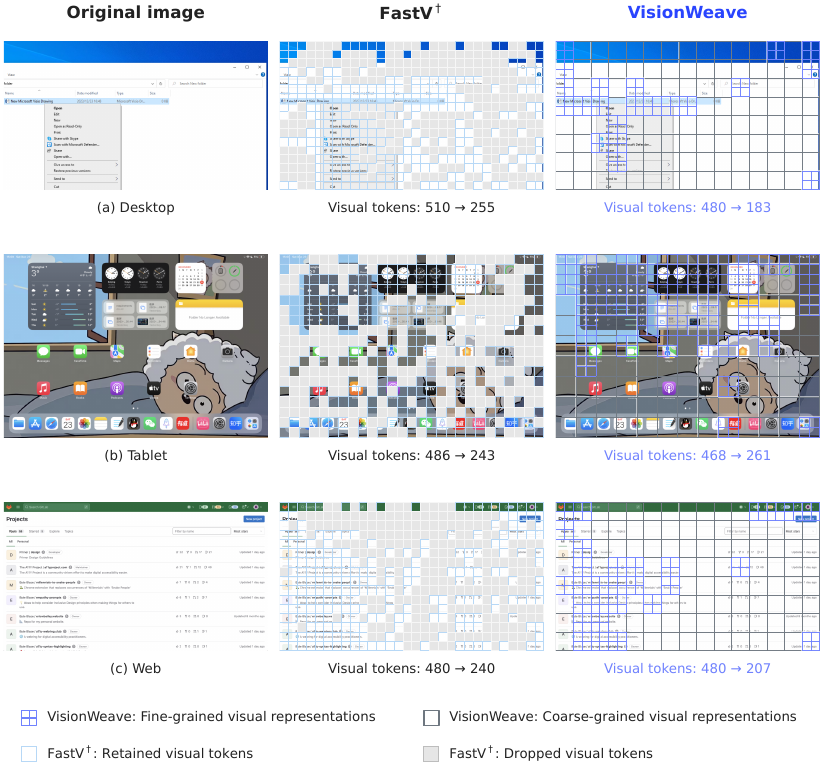}
  \caption{\method{} retains full spatial coverage through mixed-granularity
  representations, whereas FastV$^\dagger$ drops visual tokens.}
  \label{fig:app-gui-coverage}
\end{figure}

  Figure~\ref{fig:app-gui-coverage} compares visual token allocation on
  desktop, tablet, and web interfaces using Qwen3.8-27B at a 512-token
  input budget. \method{} interleaves fine- and coarse-grained
  representations, keeping every region represented within a shared
  spatial frame. In contrast, FastV$^\dagger$ drops tokens covering
  critical click targets, such as the star button in the web interface,
  leaving gaps in token coverage that can hinder accurate grounding.
   This spatially coherent
  coverage helps explain \method{}'s substantial advantage over
  FastV$^\dagger$ and VisionZip on grounding tasks.

\subsection{Case Study in DocVQA Tasks}
\label{sec:app-docvqa-visualizations}

\begin{figure}[!ht]
  \centering
  \includegraphics[width=\linewidth]{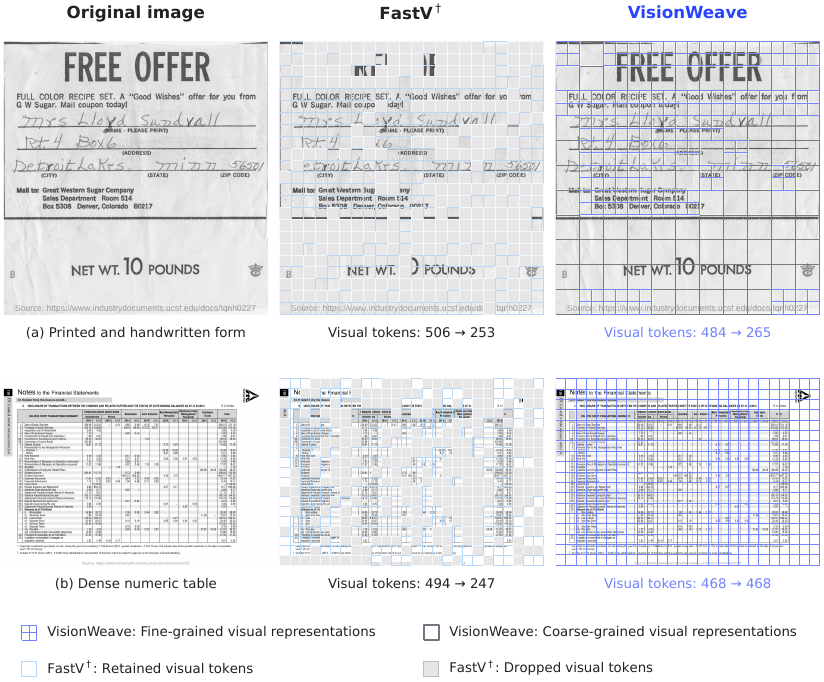}
  \caption{\method{} adapts visual granularity to document content rather
  than imposing fixed token savings.}
  \label{fig:app-docvqa-coverage}
\end{figure}

Figure~\ref{fig:app-docvqa-coverage} compares token allocation on two
DocVQA examples using Qwen3.8-27B at a 512-token input budget.
At the default threshold $t=0.5$, \method{} saves 45.2\% of tokens on
the form while retaining all native fine-grained tokens for the dense
table. FastV$^\dagger$ instead removes 50\% of tokens in both cases,
including tokens covering text. 
Even in example (a), FastV$^\dagger$ drops tokens covering informative text,
  leaving some characters only partially represented. In contrast,
  \method{} uses coarse-grained representations for larger text while
  maintaining full spatial coverage. This also
  helps explain the robustness of VisionWeave across diverse inputs.

\section{More Visualizations of Elastic Visual Weaving }
\label{sec:more_visualizations}

\begin{figure}[!t]
  \centering
  \includegraphics[width=\linewidth]{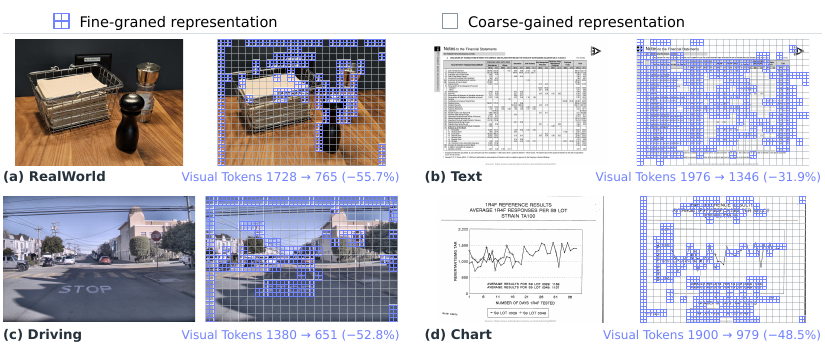}
  \caption{Elastic visual representations on Qwen3.5-4B at a 2048-token input budget.}
  \label{fig:router-teaser-4b}
\end{figure}

\begin{figure}[!t]
  \centering
  \includegraphics[width=\linewidth]{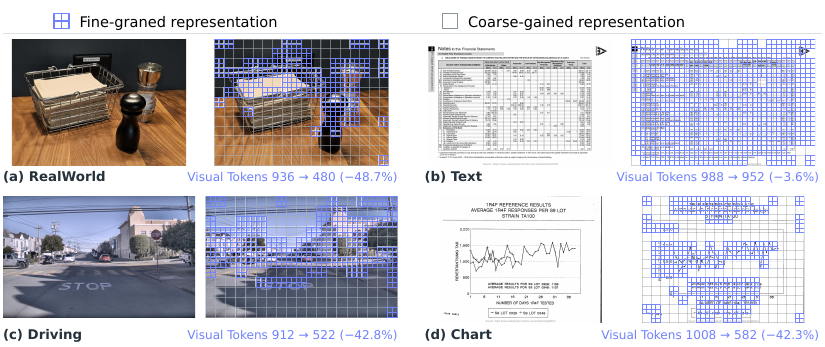}
  \caption{Elastic visual representations on Qwen3.8-27B at a 1024-token input budget.}
  \label{fig:router-teaser-27b-1024-preview}
\end{figure}

\begin{figure}[!t]
  \centering
  \includegraphics[width=\linewidth]{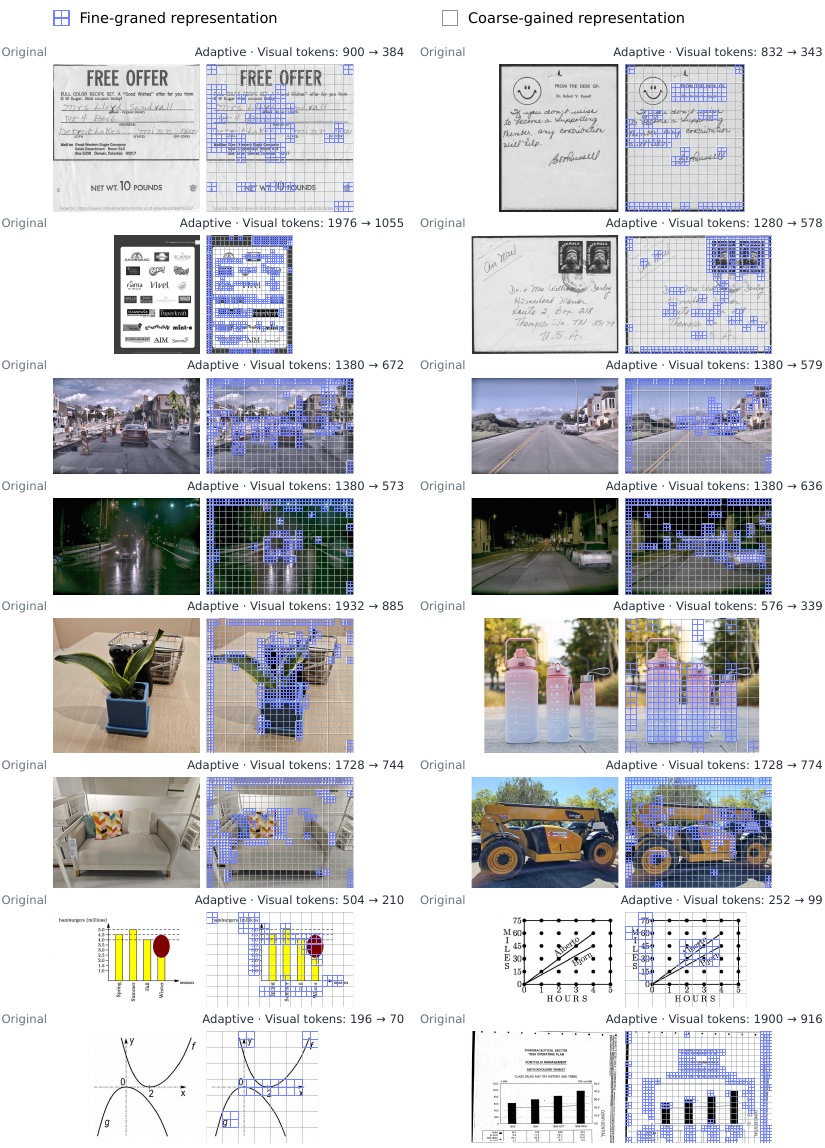}
  \caption{Additional Qwen3.5-4B visualizations at a 2048-token input budget.}
  \label{fig:more_visualizations}
\end{figure}

\begin{figure}[!t]
  \centering
  \includegraphics[width=\linewidth]{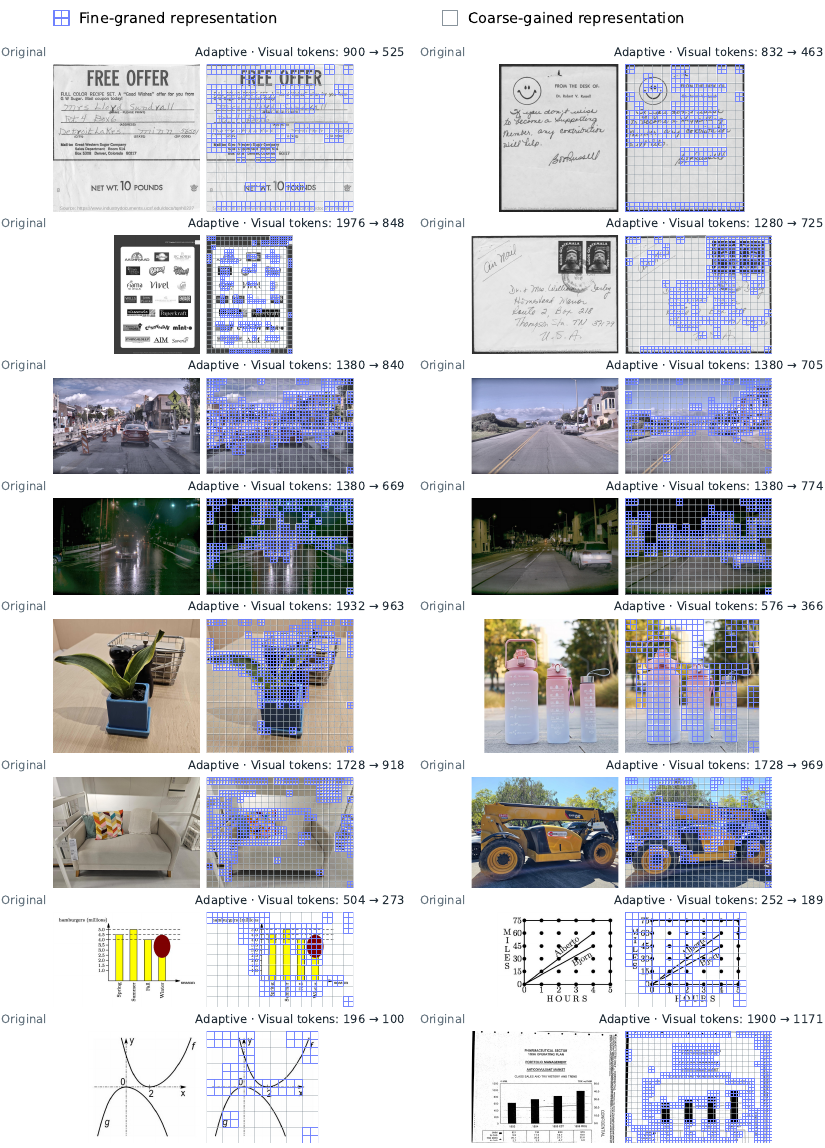}
  \caption{Qwen3.8-27B visualizations corresponding to Figure~\ref{fig:more_visualizations},
  with the same 2048-token input budget.}
  \label{fig:more-visualizations-27b}
\end{figure}

Figures~\ref{fig:router-teaser-4b}--\ref{fig:more-visualizations-27b}
show additional examples of elastic visual representations on
Qwen3.5-4B and Qwen3.8-27B, illustrating how fine- and coarse-grained
representations are interleaved according to visual content.

\section{Related Work}

\paragraph{Visual token pruning and merging.}
Visual token pruning and merging shorten visual sequences through
  heuristic or learned policies~\citep{Jeddi2025SimilarityAwareTP, Ye2024ATPLLaVAAT,chen2026variationawarevisiontokendropping,han2025filtercorrelatecompresstrainingfree,Zhang2025BeyondAO}, as exemplified by FastV and VisionZip~\citep{Chen2024AnII,Yang2024VisionZipLI}.
  Direct pruning removes tokens deemed unimportant by the selection policy
  and can substantially alter the visual context available to a pretrained
  MLLM. Our evaluations of FastV$^\dagger$ and VisionZip show modest
  performance losses on video tasks but substantial degradation on grounding
  and information-dense inputs.  Most methods also rely on preset compression ratios that cannot accommodate the varying information density of visual inputs.  In addition, despite extensive research, visual token-pruning methods rarely provide
  implementations in modern serving engines such as SGLang or vLLM,
  leaving their practical efficiency gains insufficiently validated.
  Common algorithmic choices further hinder deployment: some methods
  require explicit attention weights, which standard FlashAttention
  kernels do not expose, while others prune tokens within the LLM,
  complicating engine integration and support for dynamic batching and
  chunked prefill. Consequently, the growing body of token-pruning
  research has yet to translate into widespread adoption in practical
  LLM serving. In contrast, \method{} improves visual efficiency through
  content-adaptive selection of fine- and coarse-grained representations,
  using the same representation construction and routing during final-stage
  training and inference. Extensive evaluations on SGLang engine across tasks, input
  resolutions, and video frame budgets demonstrate consistently favorable
  efficiency--quality trade-offs, supporting elastic visual weaving as
  a robust native capability of MLLMs.

\begin{figure}[!t]
  \centering
  \includegraphics[width=\linewidth]{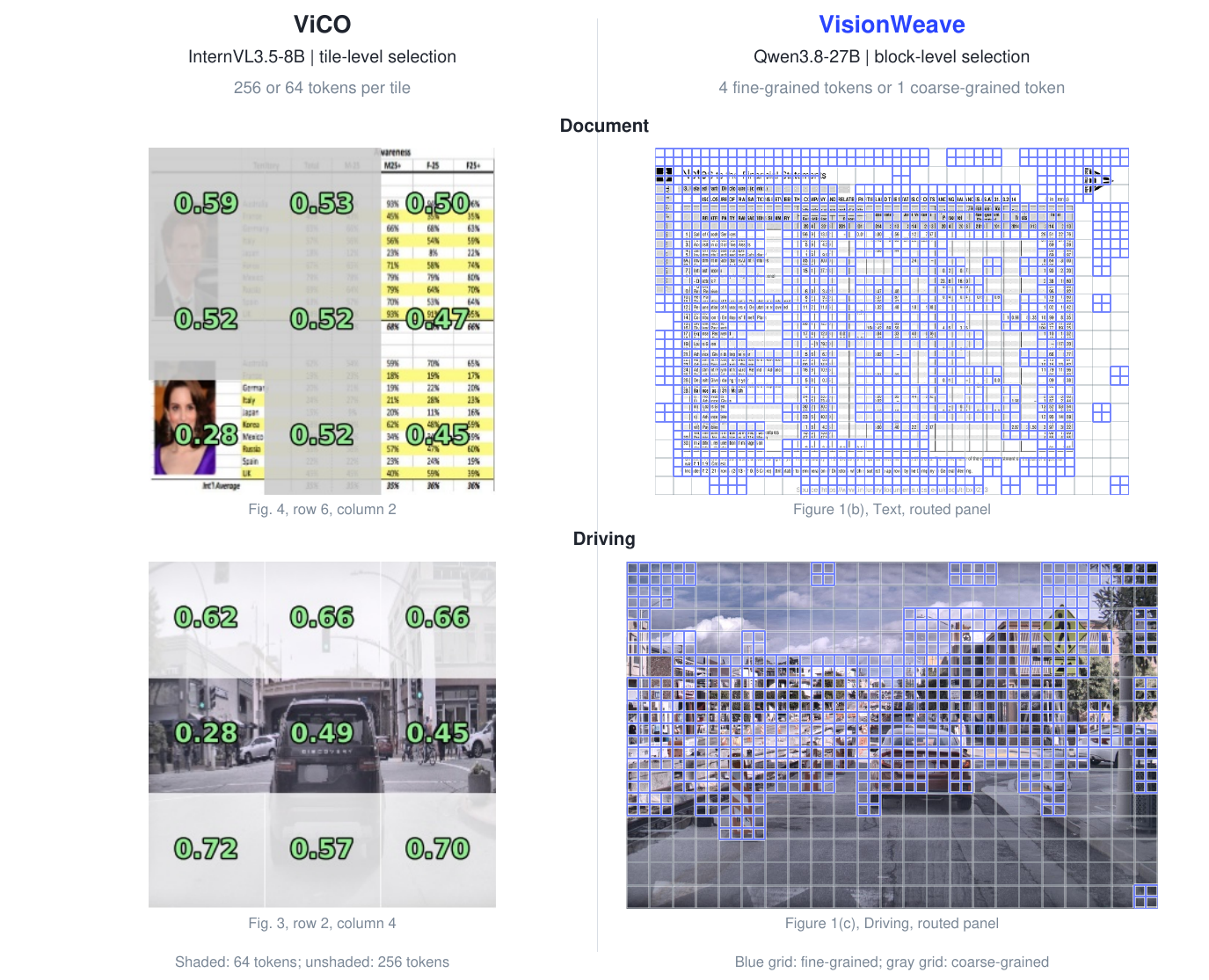}
  \caption{\method{} allocates visual granularity at a finer spatial scale than tile-level routing.
  ViCO panels are cropped from \citet{Cui2025ViCOAT}}
  \label{fig:app-vico-granularity}
\end{figure}

  \paragraph{Content-adaptive visual representations.}
  At the vision-encoder level, prior work varies patch sizes using
  low-level image statistics, such as edge density and entropy~\citep{Yu2025DynamicGM,Choudhury2025AcceleratingVT}. These approaches primarily target
  standalone vision tasks rather than learning granularity for diverse
  downstream multimodal objectives. At the MLLM level, Matryoshka Multimodal Models~\citep{cai2024matryoshkamultimodalmodels}, Matryoshka Query
  Transformer~\citep{hu2024matryoshkaquerytransformerlarge} and PARCEL~\citep{kuzucu2026parcelpoolanchoredresamplingconditioned} support multiple token budgets,
  but require the budget to be specified externally.
  ViCO~\citep{Cui2025ViCOAT} goes further by adaptively assigning a visual
  token resolution to each image tile. However, its tile-based design
  limits direct applicability to frontier MLLMs such as Qwen and Kimi,
  which use native-resolution visual processing.
  Moreover, interleaved detail-rich regions and low-information
  backgrounds call for finer spatial allocation than a single resolution
  per tile provides (Figure~\ref{fig:app-vico-granularity}).
  \method{} instead learns block-wise allocation of fine- and
  coarse-grained representations within the native-resolution pipeline.
  Our contribution is to establish this content-adaptive visual
  processing as a native capability of frontier MLLMs, combining
  autonomous token savings, robust efficiency--quality trade-offs,
  and compatibility with modern serving infrastructure.

  \paragraph{KV-cache compression.}
  KV-cache compression addresses a related but distinct aspect of
  inference efficiency. Visual token efficiency shortens the sequence
  processed by the LLM, reducing both attention and feed-forward
  computation during prefill, as well as KV-cache memory, thereby
  enabling faster time to first token~\cite{Feng2024AdaKVOK,Feng2025CriticalKVOK,Feng2025TamingTF,Liu2025MixingIW,Wang2025SparseMMHS}. In contrast, KV-cache compression
  compacts the stored key--value states without shortening the input
  token sequence. When applied after prefill, it primarily reduces
  memory usage and decoding bandwidth rather than prompt-processing
  computation. The distinction is therefore between processing fewer
  tokens and retaining more compact states for those tokens.
  Learned KV compression, exemplified by
  DeepSeek-V4~\citep{DeepSeekAI2026DeepSeekV4TH}, extends this direction
  by integrating compression into attention itself.
  Combining such mechanisms with elastic visual representation weaving
  is a promising direction: fewer visual tokens would reduce computation
  throughout the LLM, while learned KV compression could further lower
  the cache memory footprint of dense- or sparse-attention layers.

\label{app:related}

\section{Additional Experimental Details}
\label{sec:exp-details}

\subsection{Ablation of Mixed-Granularity Self-Distillation}
\label{sec:app-stage3-ablation}

We compare Qwen3.8-27B before and after Stage~3
(Table~\ref{tab:app-stage3-ablation}). Stage~3 improves seven of
eight benchmarks and raises the average from $73.15$ to $76.73$.
ScreenSpotV2 shows the largest gain ($66.59$ to $91.43$); excluding
it, the average still improves by $0.54$ points, despite a $1.18$-point
decrease on RealWorldQA. This supports self-distillation on the
hard-routed mixed-granularity sequences used at inference.

\begin{table}[!htbp]
\caption{Stage~3 improves Qwen3.8-27B performance on seven of eight benchmarks.}
\label{tab:app-stage3-ablation}
\centering
\small
\setlength{\tabcolsep}{6pt}
\renewcommand{\arraystretch}{1.05}
\begin{tabular*}{\linewidth}{@{\extracolsep{\fill}}lrrr@{}}
\toprule
Benchmark & Without Stage~3 & With Stage~3 & $\Delta$ (points) \\
\midrule
RealWorldQA & \textbf{79.08} & 77.91 & $-1.18$ \\
DocVQA & 90.76 & \textbf{91.33} & $+0.56$ \\
InfoVQA & 70.66 & \textbf{71.61} & $+0.96$ \\
ScreenSpotV2 & 66.59 & \textbf{91.43} & $+24.84$ \\
HallusionBench & 72.10 & \textbf{73.60} & $+1.51$ \\
VideoOCR & 69.85 & \textbf{70.51} & $+0.67$ \\
Video-MME & 70.44 & \textbf{71.04} & $+0.59$ \\
LongVideoBench & 65.74 & \textbf{66.42} & $+0.67$ \\
\midrule
\textit{Average} & 73.15 & \textbf{76.73} & $+3.58$ \\
\bottomrule
\end{tabular*}
\end{table}

\subsection{Ablation of Distillation Policy}
\label{sec:app-policy-ablation}
\label{sec:exp-policy-ablation}

\begin{figure}[!htbp]
\centering
\includegraphics[width=\linewidth]{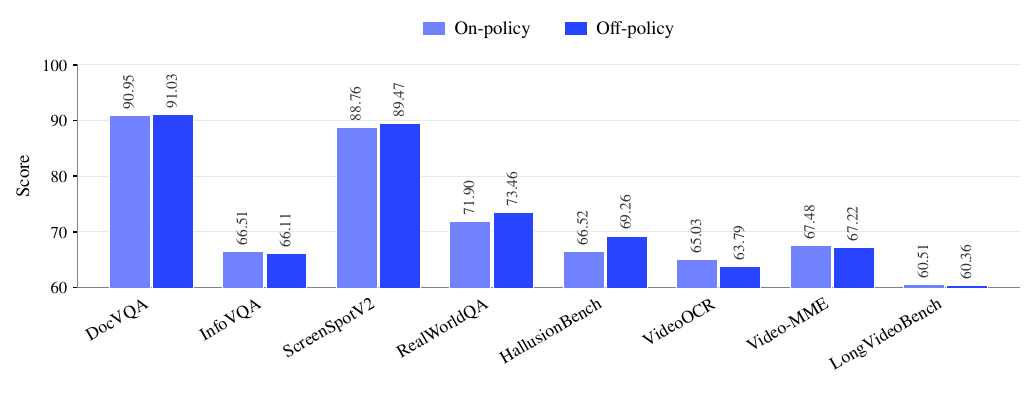}
\caption{On- and off-policy self-distillation yield comparable performance on Qwen3.5-4B.}
\label{fig:exp-policy-ablation}
\end{figure}

We also compare on- and off-policy self-distillation in Stage~3 of
Qwen3.5-4B (Figure~\ref{fig:exp-policy-ablation}). The two variants
achieve similar mean scores across eight benchmarks, each leading
on four. We therefore use off-policy self-distillation for training
efficiency.

% \subsection{Evaluation Details}
% \label{sec:app-evaluation}

% \paragraph{Task metrics.}

\begin{table}[!t]
\caption{Performance at matched dataset-level token savings.}
\label{tab:exp-matched}
\centering
\fontsize{8}{9.5}\selectfont
\setlength{\tabcolsep}{2.5pt}
\renewcommand{\arraystretch}{1.05}
\setlength{\dashlinedash}{1.2pt}
\setlength{\dashlinegap}{2pt}
\setlength{\arrayrulewidth}{0.3pt}
\begin{tabular*}{\linewidth}{@{\extracolsep{\fill}}lr:rrr@{}}
\toprule
& \multicolumn{1}{c}{Upper bound} & \multicolumn{3}{:c@{}}{Matched savings} \\
\cmidrule(lr){2-2}\cmidrule(l){3-5}
\textbf{Qwen3.5-4B} & 512 tokens & FastV$^\dagger$ & VisionZip & \method{} \\
\midrule
RealWorldQA & 74.64 & 71.76 & \underline{72.42} & \textbf{73.46} \\
DocVQA & 91.92 & \textbf{91.16} & \underline{91.09} & 91.03 \\
InfoVQA & 67.44 & \textbf{66.66} & 65.68 & \underline{66.11} \\
ScreenSpotV2 & 89.86 & \underline{87.34} & 86.08 & \textbf{89.47} \\
HallusionBench & 69.88 & 69.09 & \textbf{69.44} & \underline{69.26} \\
VideoOCR & 64.51 & \underline{63.59} & 63.28 & \textbf{63.79} \\
Video-MME & 67.78 & \textbf{67.56} & \underline{67.48} & 67.22 \\
LongVideoBench & 59.99 & \textbf{60.43} & 59.76 & \underline{60.36} \\
\midrule
\textit{Average} & 73.25 & \underline{72.20} & 71.90 & \textbf{72.59} \\
\midrule
\textbf{Qwen3.8-27B} & & & & \\
\midrule
RealWorldQA & 78.82 & \underline{78.17} & \textbf{78.43} & 77.91 \\
DocVQA & 92.47 & 88.26 & \underline{88.51} & \textbf{91.33} \\
InfoVQA & 74.13 & \underline{71.29} & 71.07 & \textbf{71.61} \\
ScreenSpotV2 & 94.42 & 43.79 & \underline{45.68} & \textbf{91.43} \\
HallusionBench & 73.25 & \underline{72.54} & \underline{72.54} & \textbf{73.60} \\
VideoOCR & 70.72 & \underline{69.79} & 69.44 & \textbf{70.51} \\
Video-MME & 70.89 & \textbf{71.15} & \textbf{71.15} & \underline{71.04} \\
LongVideoBench & 66.49 & \underline{65.82} & \underline{65.82} & \textbf{66.42} \\
\midrule
\textit{Average} & 77.65 & 70.10 & \underline{70.33} & \textbf{76.73} \\
\bottomrule
\end{tabular*}
\end{table}

\subsection{Performance at Matched Token Savings}
\label{sec:exp-matched}

Table~\ref{tab:exp-matched} gives the per-task results for the
matched-savings comparison in Section~\ref{sec:exp-main}. FastV$^\dagger$
and VisionZip match \method{}'s dataset-level savings at $t=0.5$ for
each backbone and dataset. ScreenSpotV2 contributes substantially to
the 27B average gap; excluding it, mean scores remain 74.63 for
\method{} versus 73.86/73.85 for FastV$^\dagger$/VisionZip.

\subsection{SGLang Serving Configuration}
\label{sec:app-serving}
The experiments in Section~\ref{sec:exp-serving} use our SGLang integration,
Python 3.12.3, PyTorch 2.13.0+cu129, Transformers 5.12.1, and Triton 3.7.1.
Each model runs on two A100-SXM4-80GB GPUs with LLM tensor parallelism
of two and one TP=1 vision encoder per GPU, providing encoder data
parallelism of two. Both models use BF16, Triton full/linear/ViT
attention, and NCCL with custom all-reduce disabled. The context limit
is 147,456 tokens, with 4,096-token prefill chunks. Static memory
fractions are 0.55 for the language service and 0.06 for each encoder.
FCFS and overlap scheduling are enabled; mixed prefill/decode chunks
are disabled. Padded decode CUDA graphs capture batch sizes 1, 2 and 4; prefill graphs are disabled. Client concurrency is capped at four outstanding requests.

\end{document}